\documentclass[10pt,twocolumn,letterpaper]{article}

\usepackage[pagenumbers]{cvpr} % To force page numbers, e.g. for an arXiv version

\usepackage{microtype}

\usepackage{bm}
\usepackage{bbm}
\usepackage{mathtools}
\usepackage{multirow}
\usepackage{colortbl}
\usepackage{array}
\usepackage{tabularx}
\usepackage{makecell}
\usepackage{algorithm}
\usepackage{algpseudocode}
\usepackage{placeins}
\usepackage{fontawesome5}
\usepackage{pifont}
\newcommand{\cmark}{\ding{51}}
\newcommand{\xmark}{\ding{55}}

\definecolor{bestgreen}{HTML}{81C784}
\definecolor{secondgreen}{HTML}{C8E6C9}

\definecolor{lat1}{HTML}{FFFFFF}
\definecolor{lat2}{HTML}{F3F8FD}
\definecolor{lat4}{HTML}{E4EFF9}
\definecolor{lat8}{HTML}{D3E5F4}
\definecolor{lat16}{HTML}{BFD8EE}
\definecolor{lat32}{HTML}{A9CAE7}
\definecolor{lat64}{HTML}{92BCE0}
\newcommand{\prodcell}[3]{% #1=L, #2=N, #3=content
  \begingroup
  \count0=\numexpr#1*#2\relax
  \ifnum\count0=1 \cellcolor{lat1}\fi
  \ifnum\count0=2 \cellcolor{lat2}\fi
  \ifnum\count0=4 \cellcolor{lat4}\fi
  \ifnum\count0=8 \cellcolor{lat8}\fi
  \ifnum\count0=16 \cellcolor{lat16}\fi
  \ifnum\count0=32 \cellcolor{lat32}\fi
  \ifnum\count0=64 \cellcolor{lat64}\fi
  #3%
  \endgroup
}

\definecolor{cvprblue}{rgb}{0.21,0.49,0.74}
\usepackage[pagebackref,breaklinks,colorlinks,allcolors=cvprblue]{hyperref}
\makeatletter
\providecommand{\theHALG@line}{}
\renewcommand{\theHALG@line}{\thealgorithm.\arabic{ALG@line}}
\makeatother

\definecolor{MyGreen}{rgb}{0.02,0.5,0.02}
\definecolor{myRed}{rgb}{0.8, .2, .2}
\definecolor{myBlack}{rgb}{0.0, 0.0, 0.0}

\def\confName{Arxiv}
\def\confYear{2027}

\makeatletter
\patchcmd{\@maketitle}{\vskip .375in}{\vskip 0in}{}{\PackageWarning{main}{Failed to patch title spacing}}
\patchcmd{\endabstract}{\vspace*{12pt}}{\vspace{12pt}}{}{\PackageWarning{main}{Failed to patch abstract spacing}}
\renewcommand\paragraph{\@startsection{paragraph}{4}{\z@}%
  {4pt}%
  {-1em}%
  {\normalfont\normalsize\bfseries}}
\makeatother

\newcommand{\method}{SPARC\xspace}

\title{\faCut~SPARC: Geometry-Guided Attention Pruning \\ for Long-Horizon Camera-Controlled Video Generation}

\author{
Zesong Yang$^{1}$ \quad
Weikai Chen$^{2\ddagger}$ \quad
Liyuan Cui$^{1}$ \quad
Yuan Li$^{1}$ \quad
Jiaxin Huang$^{1}$ \\
Lutao Jiang$^{3}$ \quad
Runze Zhang$^{2}$ \quad
Yingda Yin$^{2}$ \quad
Xiaoyang Huang$^{2}$ \quad
Kai Yan$^{2}$ \\
Keyang Luo$^{2}$ \quad
Xin WANG$^{2}$ \quad
Hujun Bao$^{1}$ \quad
Zhaopeng Cui$^{1\dagger}$ \\[0.5em]
$^{1}$State Key Laboratory of CAD\&CG, Zhejiang University \quad
$^{2}$LIGHTSPEED \quad
$^{3}$HKUST(GZ) \\[0.25em]
}

\begin{document}
% \maketitle

\twocolumn[{%
\renewcommand\twocolumn[1][]{#1}%
\maketitle
\begin{center}
  \centering
  \vspace{-1.5 em}
  \includegraphics[width=0.975\textwidth]{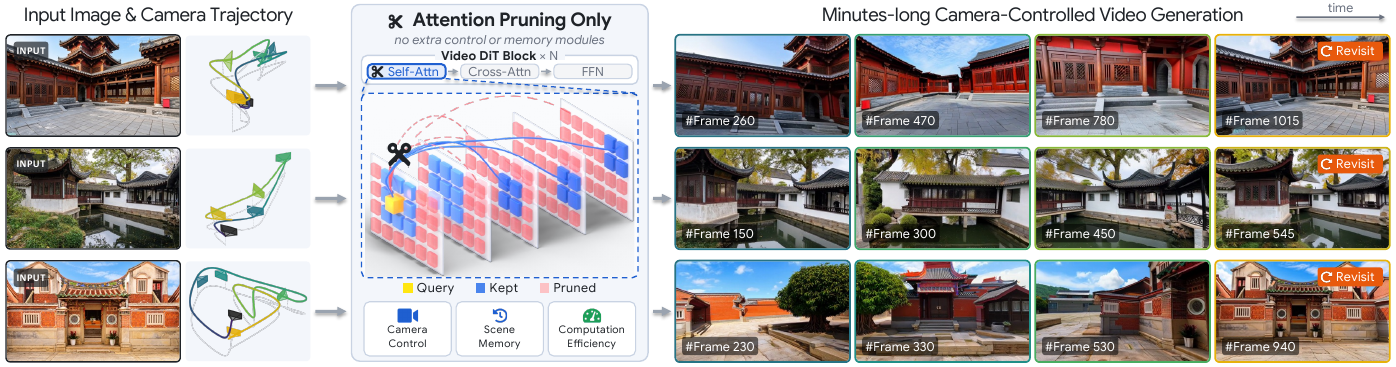}
  \captionof{figure}{
    \textbf{Less Attention, Better Control, Persistent Memory.} 
    Given a single image, \method generates minute-long, photorealistic video through attention pruning alone, achieving precise camera control and persistent scene consistency without additional learnable modules.
  }
  \vspace{1.5 em}
  \label{fig:teaser}
\end{center}
}]

\setlength{\abovedisplayskip}{0.5 em}
\setlength{\belowdisplayskip}{0.5 em}

{
\renewcommand{\thefootnote}{}
\footnotetext{$^{\ddagger}$Project Lead. \quad $^{\dagger}$Corresponding Author.}
}

\begin{abstract}
    We present \method, a geometry-guided attention pruning framework that turns the native self-attention of pretrained video diffusion transformers into a unified mechanism for camera control and persistent scene memory. 
    Our key insight is that \textbf{attention connectivity} can simultaneously serve as geometric conditioning for camera control and spatial addressing for historical memory retrieval.
    % Specifically, previously observed regions rely on geometrically corresponding historical evidence, while newly revealed regions benefit from broader contextual interactions.
    % \method exploits this asymmetry by routing observed tokens to spatially aligned historical features and local spatiotemporal context, while preserving dense attention for unobserved regions. 
    Motivated by the asymmetric contextual requirements of reconstructing observed content and synthesizing novel regions, SPARC restricts observed tokens to geometrically corresponding historical tokens and local spatiotemporal context, while preserving dense attention for unobserved regions. 
    Correspondence-aware attention biasing and order-invariant historical positioning further enable stable retrieval across long temporal gaps.
    Without introducing additional learnable control or memory modules, \method achieves precise camera adherence and faithful scene revisitation while substantially reducing attention computation.
    Extensive experiments demonstrate the effectiveness of aggressive attention pruning in improving camera adherence, revisit consistency, and visual fidelity, with an inference speedup of over 50$\times$ when combined with four-step distillation. Project page is available at \url{https://zju3dv.github.io/SPARC}.
\end{abstract}
 % ✅
\section{Introduction}
\label{sec:intro}

% * Task Motivation;
% * ---- 视频生成的发展与长程相机控制的两个核心要求：camera adherence 和 revisit consistency

Recent advances in video diffusion transformers~\cite{wan2025wan,seedance2026seedance2,klingai2026kling3,googledeepmind2025veo3,alibaba2026happyhorse,minimax2026h3} have enabled increasingly realistic and temporally coherent video generation, bringing interactive world generation closer to reality~\cite{alibaba2026happyoyster,parkerholder2025genie3, chen2026hy,zhang2026echowmopenenterableomnimodal,lingbot-world-v2}. 
However, long-horizon generation under prescribed camera trajectories remains challenging.
Beyond synthesizing plausible frames, a model must accurately follow camera movements and preserve scene identity when returning to previously visited locations. 
Existing approaches primarily rely on camera-conditioning modules~\cite{bai2025recammaster, he2024cameractrl, li2025cameras,zhang2026unified}, explicit 3D reconstruction~\cite{yu2024viewcrafter,ren2025gen3c,wu2025video,zhao2026spatia,yang2026geometry}, or historical frames as context~\cite{xiao2026worldmem,sun2025worldplay,yu2025cam,hong2025relic,shen2026lyra} to support controllable long-horizon generation. 
While effective, these mechanisms introduce additional architectural complexity or rely on computationally expensive dense attention over historical observations.

% ! 从计算冗余转向信息交互机制
Rather than adding dedicated control or memory modules, we explore an alternative perspective: 
% can the attention connectivity itself of a pretrained video diffusion transformer serve as a mechanism for camera control and persistent scene memory? 
can attention connectivity itself serve as a unified mechanism for camera control and persistent scene memory in pretrained video diffusion transformers?
Both capabilities require spatially grounded access to historical visual evidence, yet conventional dense attention allows unrestricted interactions with historical tokens, without explicitly enforcing spatial correspondence. 
Such interactions introduce computational redundancy and may interfere with accurate historical retrieval.

\begin{figure}[t]
    \centering
    \includegraphics[width=0.475\textwidth]{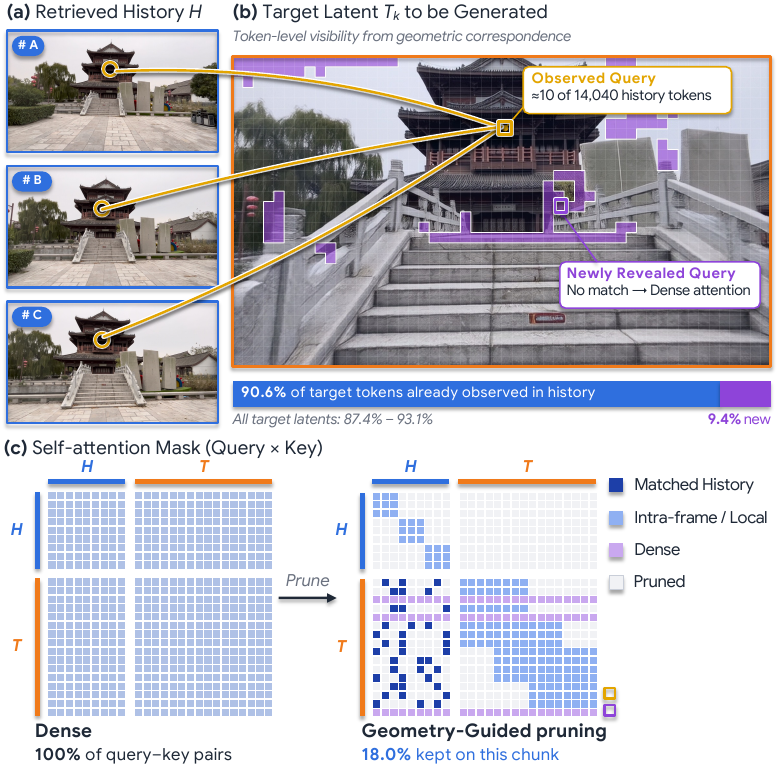}
    \caption{
    \textbf{Redundancy of dense full self-attention.}
    Most target regions have been observed in historical frames and require only sparse geometric correspondences for faithful reconstruction, while newly revealed regions benefit from broader contextual interactions.
    }
    \vspace{-0.25 em}
    \label{fig:motivation}
\end{figure}

Our key insight is that \textit{attention pruning can serve as geometric conditioning and memory addressing, rather than merely a computational optimization}. 
Given a target camera pose, geometric correspondences determine which historical tokens each target token can access.
Crucially, observed regions attend to geometrically matched historical evidence and local context, while unseen regions retain broader attention for plausible synthesis. 
As shown in Fig.~\ref{fig:motivation}, most target regions require only sparse historical correspondences, enabling efficient information routing that supports both camera control and persistent memory retrieval.

Building upon this insight, we introduce \method, a geometry-guided attention pruning framework that directly restructures the native self-attention of pretrained video diffusion transformers.
\method constructs token-level historical correspondences through local geometric projection and applies asymmetric attention routing based on target-region visibility.
Observed tokens access geometrically matched historical evidence and local temporal context, whereas unobserved tokens retain dense attention for flexible content synthesis.
Correspondence-aware attention biasing further strengthens historical evidence against the substantially larger local context, while order-invariant positioning eliminates artificial temporal ordering biases among retrieved historical observations. 
% These designs enable camera control and persistent scene consistency without additional learnable control or memory modules, while simultaneously reducing attention computation.
Without introducing additional learnable control or memory modules, \method exploits the pretrained model's inherent capabilities through lightweight self-attention adaptation, enabling camera control and persistent scene consistency.
Combined with four-step distillation, \method further achieves an inference speedup of over 50$\times$ relative to the original backbone.
% Experiments on long-horizon camera-controlled video generation demonstrate substantial improvements in camera adherence and revisitation fidelity. Combined with four-step distillation, \method{} achieves over 50$\times$ DiT inference acceleration relative to the original backbone.
% Our main contributions are:

Our main contributions are summarized as follows:
\begin{itemize}
    \item We reveal that camera control and persistent memory share a common requirement for spatially grounded historical retrieval, motivating a unified attention-based solution.

    \item We propose \method, a geometry-guided asymmetric attention pruning framework that uses native self-attention for camera conditioning and persistent memory retrieval, without additional learnable control or memory modules.

    \item Extensive experiments demonstrate that \method achieves superior camera adherence, revisit consistency, and visual fidelity, while substantially reducing attention computation and inference latency for minute-long video generation.
    % Extensive experiments demonstrate the effectiveness of \method in improving camera adherence, revisit consistency, and visual fidelity, alongside substantial reductions in attention computation and inference latency for minute-long video generation.
\end{itemize} % ✅
\section{Related Work}
\label{sec:related}

\paragraph{Camera Control in Video Generation.}
% 一些工作尝试 incorporate camera trajectories with video generation to enable interactive video generation with viewpoint changes. 其中一些工作对 camera represenetaions such as plucker rays or camera poses 编码为 condition embedding~\cite{worldmem, recammaster, cameractrl}, 注入到输入latent，或者通过修改 attention positional encoding 机制~\cite{PRoPE, UCPE, SCOPE}. 此外还有工作通过引入额外的 3D 显式先验, 通过投影得到 pixel-aligned 的 3D 条件以提供 spatially grounded 的 camera control~\cite{viewcrafter,gen3c,ex4D, OmniDirector, wonder, ucm, PE-Field}. 这些工作都引入了额外的 learnable condition modules, 增加了额外的计算和内存开销.
Recent works incorporate camera trajectories into video generation for controllable viewpoint changes. Some encode camera poses or Plücker rays as conditioning embeddings~\cite{xiao2026worldmem, bai2025recammaster, he2024cameractrl, zhou2025stable}, or modify attention mechanisms to incorporate camera geometry~\cite{li2025cameras,zhang2026unified,yin2026scope}. Others leverage explicit 3D priors to provide pixel-aligned geometric guidance through projection~\cite{yu2024viewcrafter,ren2025gen3c,hu2025ex4dextremeviewpoint4d,liu2024omnidirector,xu2026wonder,xu2026ucm,bai2025positional}. These approaches often introduce additional conditioning mechanisms or geometric representations, increasing architectural complexity and computational overhead.

\paragraph{Long-Horizon Memory in Video Generation.}
% 随着 long-horizon generation rollout的进行，越来越长的历史会造成 quadratic attention complexity and computational collapse, 因此有必要对 history 进行高效管理. 一些工作 reconstruct historical observations into 3D memories and project them to target viewpoints as pixel-aligned conditions~\citep{wu2025video,zhao2026spatia,zhang2026worldstereo,lee20253dscenepromptingsceneconsistent,wang2026anchorweave,chen2026anyrecon}. 一些工作利用camera poses的 FoV overlap 来 retrieval keyframes~\cite{worldmem,vmem,worldplay,context-as-memory}. 还有一些工作直接对历史进行压缩~\cite{GameCraft,Yume1.5,RELIC,InfiniteWorld,FramePack,Lyra2}，slows history growth but incurs severe information loss. 这些方法都需要额外的 learnable control and memory management modules, 增加了额外的计算和内存开销. In contrast, \method秉承 less is more 的理念，直接对 self-attention 进行 pruning，通过高效 retrieve historical evidence, 实现了precise camera control and long-horizon persistent scene memory without any additional learnable modules.
Long-horizon generation requires efficient history management to mitigate the quadratic complexity of growing attention contexts. Existing methods reconstruct explicit 3D memories for target-view projection~\cite{wu2025video,zhao2026spatia,zhang2026worldstereo,lee20253dscenepromptingsceneconsistent,wang2026anchorweave,chen2026anyrecon}, retrieve relevant keyframes based on camera frustum overlap~\cite{xiao2026worldmem,li2025vmem,sun2025worldplay,yu2025cam}, or compress historical context at the cost of potential information loss~\cite{li2025hunyuangamecrafthighdynamicinteractivegame,mao2025yume15textcontrolledinteractiveworld,hong2025relic,wu2026infiniteworld,zhang2025framepack,shen2026lyra}. However, these methods often incur additional memory management overhead or redundant attention computation. In contrast, \method directly prunes the backbone's native self-attention to retrieve geometrically relevant historical evidence, enabling precise camera control and persistent scene memory without additional learnable modules.

% ✅
\section{Methods}
\label{sec:methods}

% * Camera control and persistent scene memory do not necessarily require dedicated control or memory modules. Instead, they can emerge from selectively restructuring the native self-attention of a pretrained video diffusion model.

\begin{figure*}[t] % 使用 !t 尝试强制图像置顶
    \centering
    \scriptsize
    \vspace{-2.0 em}
    \includegraphics[width=0.975 \textwidth]{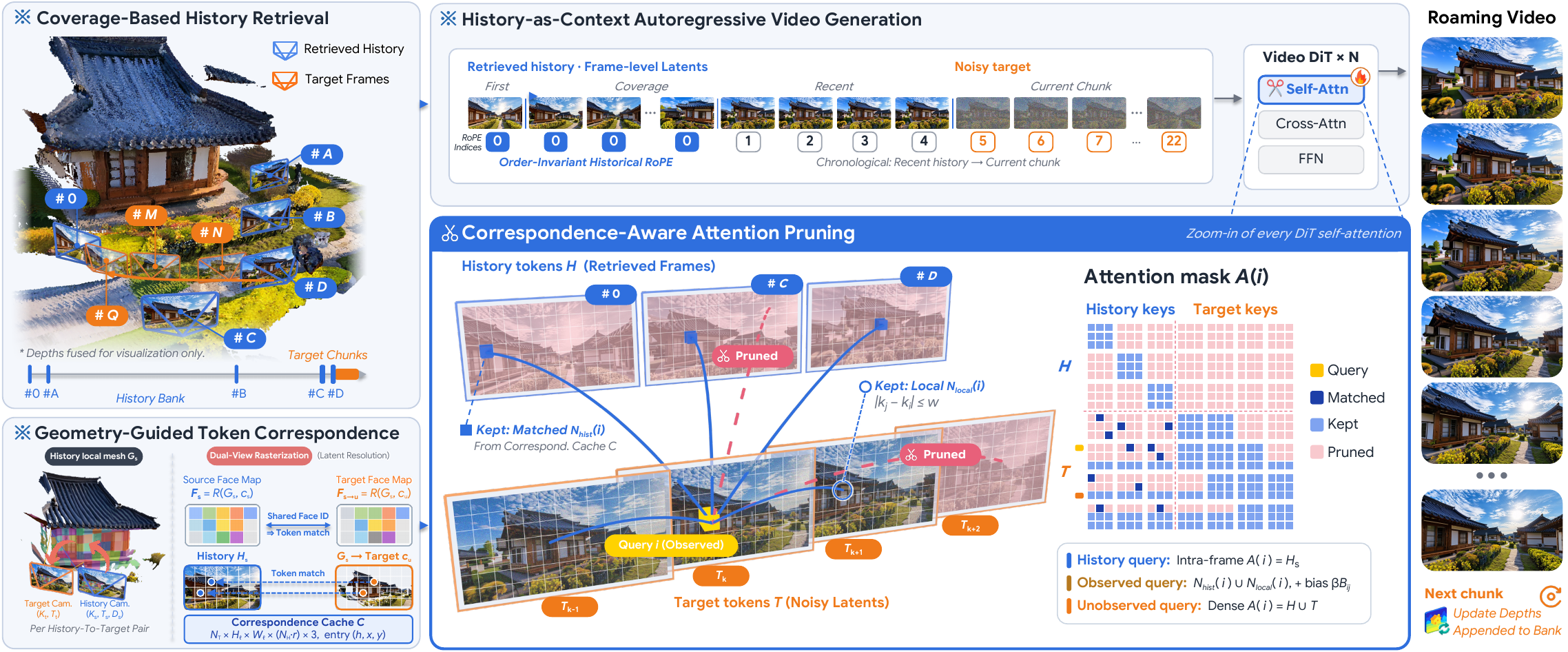}
    \caption{
    \textbf{System Overview.}
    For each target chunk, \method establishes token-level correspondences with historical observations through local geometric projection, and then prunes the backbone's native self-attention to focus on geometrically corresponding historical tokens and local spatiotemporal context for previously observed regions, while newly revealed regions retain dense attention for flexible content synthesis. Historical RoPE re-indexing is employed to eliminate artificial temporal ordering biases among retrieved historical observations. 
    The generated observations are subsequently appended to the history bank for continuous rollout.
    }
    \label{fig:pipeline}
  \end{figure*}
  
\subsection{Problem Formulation and Preliminaries}
\label{sec:method:overview}

\paragraph{Camera-Conditioned Autoregressive Video Generation.}
Given an initial image $I_1$, a text prompt $y$, and a prescribed camera trajectory $\{c_t\}_{t=1}^{T}$, our goal is to autoregressively generate a long-horizon video with precise camera control and persistent scene consistency. 
Following \cite{yu2025cam}, we maintain a streaming history bank $\mathcal{B}_t=\{(I_s,z_s,c_s,D_s)\}_{s<t}$, containing previously generated frames, their VAE latents, camera poses, and estimated depth maps.
For each upcoming camera chunk $c_{t+1:t+k}$, we retrieve a compact set of historical keyframes $\mathcal{H}_t^{\mathrm{ret}}\subseteq\mathcal{B}_t$ according to their geometric coverage of the target views, as detailed in Sec.~\ref{sec:method:retrieval_rerope}. 

We adopt a mixed-granularity latent representation: historical frames are independently encoded into frame-level latents, while target video chunks retain the pretrained VAE's temporal compression ratio of $r=4$.
The retrieved historical latents then serve as contextual memory and are concatenated with the noisy target latents along the temporal dimension.
The video generator then samples the next latent chunk as:
\begin{equation}
    z_{t+1:t+k}
    \sim
    p_\theta\!\left(
        \cdot
        \mid
        \mathcal{H}_t^{\mathrm{ret}},
        c_{t+1:t+k},
        y
    \right).
\end{equation}
The generated latents are decoded into video frames, which are subsequently re-encoded into frame-level latents and appended to the history bank together with their camera poses and estimated depths, enabling continuous generation.
% The generated latents are decoded into video frames and subsequently re-encoded into frame-level latents, which are appended to the history bank alongside camera poses and estimated depths for continuous autoregressive generation.

\paragraph{Flow-Matching Objective.}
We train the video generator with the standard flow-matching objective~\citep{lipman2022flow}.
For a target latent block $\mathbf{z}_{t+1:t+k}$, we sample Gaussian noise
$\boldsymbol{\epsilon}\sim\mathcal{N}(\mathbf{0},\mathbf{I})$ and a flow timestep
$\tau\sim\mathcal{U}(0,1)$, and construct the linear interpolation:
\begin{equation}
    \mathbf{z}^{\tau}_{t+1:t+k}
    =
    (1-\tau)\mathbf{z}_{t+1:t+k}
    +
    \tau\boldsymbol{\epsilon}.
\end{equation}
The conditional velocity field is optimized as:
\begin{equation}
\label{eq:flow_matching}
\begin{aligned}
    \mathcal{L}_{\mathrm{FM}}
    =
    \mathbb{E}_{\mathbf{z},\boldsymbol{\epsilon},\tau}
    \Big[
        \big\|
            & v_{\theta}
            \left(
                \mathbf{z}^{\tau}_{t+1:t+k},
                \tau
                \mid
                \mathcal{H}_t,
                \mathbf{c}_{t+1:t+k}, y
            \right)
            \\
            & -
            \left(
                \boldsymbol{\epsilon}
                -
                \mathbf{z}_{t+1:t+k}
            \right)
        \big\|_2^2
    \Big].
\end{aligned}
\end{equation}

Our central objective is to enable camera control and persistent memory retrieval by directly restructuring the backbone's native self-attention. To this end, we first establish geometric correspondences between historical observations and target latent tokens, and subsequently use them to guide aggressive attention pruning.
% \vspace{-2.0 em}
\subsection{Geometry-Guided Token Correspondence}
\label{sec:method:correspondence}

% * 明确历史信息在哪里 -> Geometry determines where to attend.
% To identify relevant historical evidence for each target token, we establish explicit token-level correspondences through geometric projection. Rather than reconstructing a globally fused 3D scene, we construct independent local geometry for each historical observation, allowing geometric correspondences to be established without maintaining a persistent 3D appearance representation.
Our attention pruning requires identifying which historical tokens provide spatial evidence for each target token. Rather than constructing a globally fused 3D scene, we establish these token-level correspondences directly from per-frame local geometry with geometric projection.
% TODO: 强调 global fusion 会导致误差累积

\paragraph{Local Geometric Anchors.}
For each retrieved historical

\noindent frame $s$, we back-project its estimated depth map $D_s$  with corresponding camera intrinsics $K_s$ and extrinsics $T_s$, and connect neighboring pixels according to the image-grid topology, obtaining a local triangular mesh $\mathcal{G}_s=(\mathcal{V}_s,\mathcal{F}_s)$.
% , where $V_s$ and $F_s$ denote the mesh vertices and faces, respectively.
Since modern video VAEs aggressively compress the spatial resolution,
we construct the mesh at the original image resolution to preserve geometric discontinuities that would otherwise be blurred through spatial latent compression.

\paragraph{History-to-Target Token Association.}
We then rasterize $\mathcal G_s$ under the historical camera $c_s$ and a target camera $c_u$ at the latent spatial resolution $H_\ell\times W_\ell$, obtaining face-ID maps for establishing token correspondence as shown in Fig.~\ref{fig:pipeline}:
\begin{equation}
    F_s = \mathcal{R}(\mathcal{G}_s,c_s), \quad
    F_{s\rightarrow u} = \mathcal{R}(\mathcal{G}_s,c_u),
\end{equation}
where each valid entry records the ID of the visible mesh face associated with a latent patch token. Since both maps originate from the same local mesh, shared triangle IDs establish direct source--target token associations:
\begin{equation}
    \mathcal{C}_{s\rightarrow u}
    =
    \left\{
    (i,j)\mid
    F_{s\rightarrow u}(i)=F_s(j)>0
    \right\},
\end{equation}
where $i$ and $j$ denote target and historical spatial token indices, respectively. 
Given the distinct temporal representations of historical and target latents as described in Sec.~\ref{sec:method:overview}, 
one target latent time step $k$ aggregates visual information from multiple consecutive frames $\mathcal W_k$.
We therefore pool historical correspondence candidates across these frames and all retrieved observations:
\begin{equation}
    \mathcal N_{\mathrm{hist}}(k,i)
    =\bigcup_{s\in\mathcal H_t^{\mathrm{ret}}}
      \ \bigcup_{u\in\mathcal W_k}
      \big\{(s,j)\mid(i,j)\in\mathcal C_{s\rightarrow u}\big\}.
    \label{eq:historical_correspondence}
\end{equation}
We organize the resulting historical token indices as a fixed-size correspondence cache:
$
    \mathbf{C}\in
    \mathbb{Z}^{N_T\times H_\ell\times W_\ell
    \times(N_H \cdot r)\times3},
$
where $N_T$ and $N_H$ denote the numbers of target latent frames and retrieved historical frames. Each entry stores $(h,x,y)$, specifying the historical frame index and spatial coordinates of a corresponding source token, with unmatched entries marked as invalid.
Crucially, geometry is used to determine \emph{where the pretrained model attends}, rather than to provide a separate pixel-aligned conditioning pathway.

\subsection{Correspondence-Aware Attention Pruning}
\label{sec:method:attention_prune}

Dense self-attention uniformly models interactions among 
historical and target tokens, despite their fundamentally different contextual requirements. Previously observed regions primarily require geometrically relevant historical evidence for faithful recovery, whereas newly revealed regions depend on broader contextual information for plausible synthesis. Motivated by this asymmetry, we introduce a correspondence-aware attention pruning strategy that selectively preserves essential interactions within the original self-attention architecture.

\paragraph{Asymmetric Attention Routing.}
Let $\mathcal H_s$ denote the tokens of retrieved historical frame $s$, $\mathcal H=\bigcup_s\mathcal H_s$ the complete retrieved history, and $\mathcal T$ the target tokens. For a target token $i$ at latent time step $k_i$, let $\mathcal N_{\mathrm{hist}}(i)$ denote the correspondence set in Eq.~\eqref{eq:historical_correspondence}. We mark it as observed when at least one valid historical match exists, i.e., $m_i^{\mathrm{vis}}=\mathbf 1[|\mathcal N_{\mathrm{hist}}(i)|>0]$. 
These historical correspondences provide strong spatial constraints but are insufficient to maintain coherence among newly generated regions.
% To preserve local coherence, we additionally relax the attention sparsity constraint with local spatiotemporal attention and define a temporal neighborhood:
To preserve local coherence, we additionally relax the attention sparsity constraint with local spatiotemporal attention and define a temporal neighborhood:
\begin{equation}
    \mathcal N_{\mathrm{local}}(i)
    =\big\{j\in\mathcal T\mid |k_j-k_i|\le w\big\},
    \label{eq:local_neighborhood}
\end{equation}
where $w$ is the temporal window radius.
% spatial attention is unrestricted within the selected target latent time steps. 
The set of accessible keys for each query token is then:
\begin{equation}
    \mathcal A(i)=
    \begin{cases}
        \mathcal H_s,
        & i\in\mathcal H_s,\\
        \mathcal N_{\mathrm{hist}}(i)\cup\mathcal N_{\mathrm{local}}(i),
        & i\in\mathcal T,\ m_i^{\mathrm{vis}}=1,\\
        \mathcal H\cup\mathcal T,
        & i\in\mathcal T,\ m_i^{\mathrm{vis}}=0.
    \end{cases}
    \label{eq:asymmetric_attention}
\end{equation}
The first case limits clean historical queries to intra-frame attention, avoiding interactions with noisy target features. The second case preserves geometry-aligned historical retrieval together with local spatiotemporal refinement, supporting precise camera control, revisit consistency and coherent generation. The third case leaves the pretrained model's full-context attention intact, with broader contextual information for plausible scene completion.
% ! 需要强调的是，camera information is conveyed through attention connectivity rather than additional pose embeddings.

\paragraph{Correspondence-Prioritized Attention.}
For observed target regions, geometrically corresponding historical tokens and local target tokens are normalized within the same attention softmax. Since local target context contains substantially more tokens, model can become biased toward nearby generated content, weakening the influence of historical evidence.
We address this imbalance with an additive bias to the attention logits associated with valid historical correspondences:
% To ensure that geometrically corresponding historical information remains influential, we introduce correspondence-prioritized attention biasing. Specifically, we assign an additive bias to the attention logits associated with valid historical correspondences:
\begin{equation}
    o_i = \sum_{j\in\mathcal A(i)} \operatorname{softmax}_j\Big(\frac{q_i k_j^\top}{\sqrt{d}} + \beta B_{ij}\Big)\, v_j .
    \label{eq:sparse_attention}
\end{equation}
Here, $o_i$ is the attention output of query token $i$, computed only over the keys and values in its allowed scope $\mathcal A(i)$. The correspondence bias $B_{ij}=\mathbf 1[j\in\mathcal N_{\mathrm{hist}}(i)]$ marks geometrically matched historical keys, with $\beta>0$ controlling their relative importance. This bias strengthens the contribution of geometry-addressed historical features while retaining content-adaptive interactions with neighboring target tokens.

\paragraph{Attention Complexity.}
Let $N_H$ be the number of retrieved historical frames, $N_T$ the number of target latents, and $P=H_\ell \cdot W_\ell$ the number of spatial tokens per latent. Conventional dense attention over the concatenated sequence of $N=N_H+N_T$ latents requires $\mathcal O(N^2P^2)$ pairwise interactions. Let $\rho$ be the fraction of observed target tokens, $\bar{k}$ their average number of matched historical keys, and $L_w\le 2w+1$ the number of target time steps in a local neighborhood. Our asymmetric attention pattern has complexity:
\begin{equation}
    \mathcal O\big(N_HP^2 + N_TP\,[\rho(\bar{k}+L_wP) + (1-\rho)NP]\big).
    \label{eq:attention_complexity}
\end{equation}
Thus, the computational benefit grows with the fraction of geometrically observed target regions, while dense modeling capacity is preserved for novel views. Consequently, the majority of target queries can avoid full-context attention and instead access only a small set of geometrically relevant historical tokens and local target context.

\begin{figure*}[t]
    \centering
    \vspace{-0.25 em}
    \includegraphics[width=0.975\textwidth]{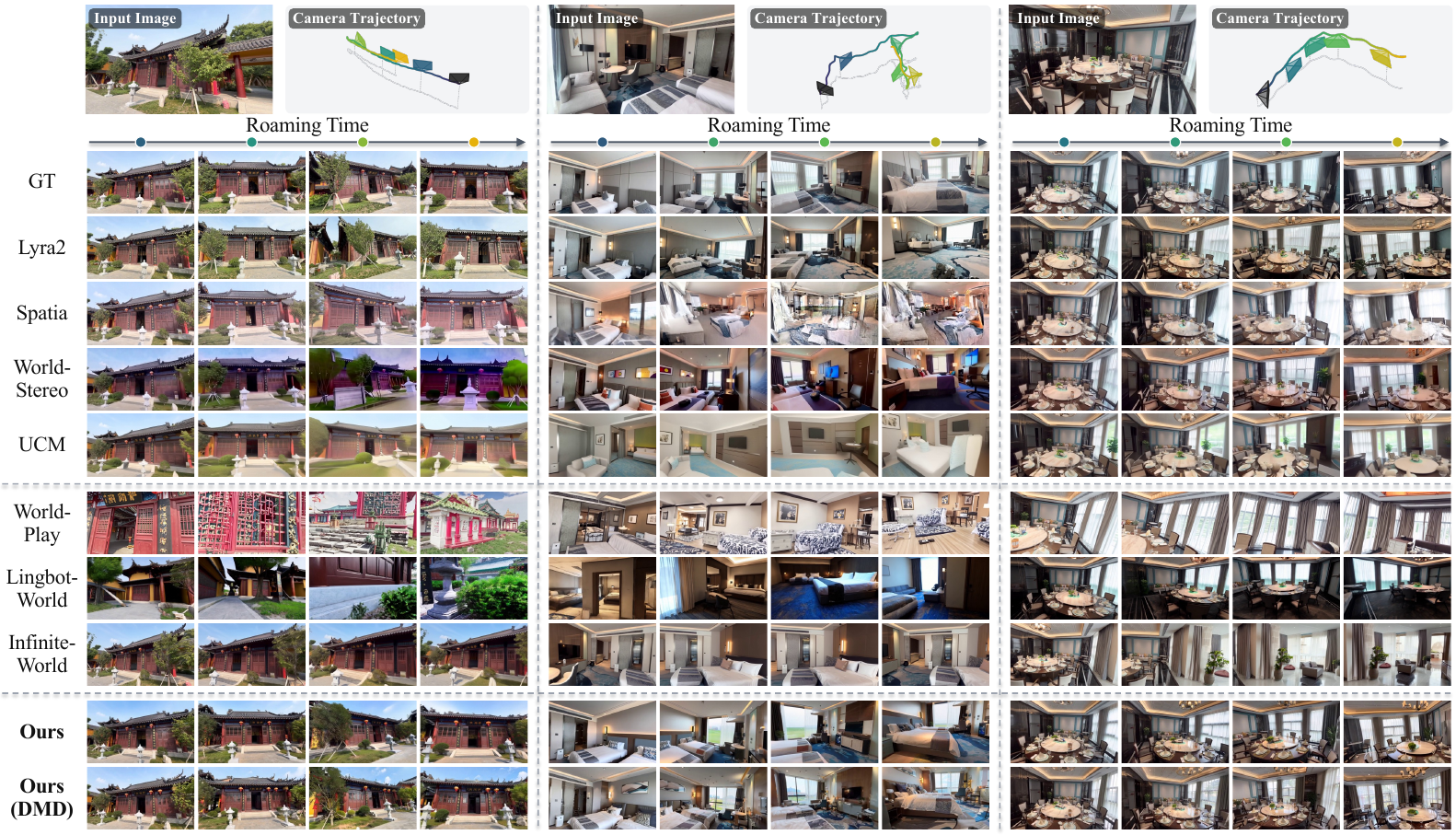}
    \caption{
    \textbf{Qualitative comparison on DL3DV-Eval.}
    Given the same input image and camera trajectory, we show frames generated with different methods along the roaming time.
    Explicit-3D baselines drift from the target trajectory or accumulate geometric distortions, and implicit baselines fail to follow the prescribed camera path, whereas \method preserves scene appearance and accurate camera control throughout the rollout. \textbf{See Supp. for additional scenes and video comparisons.}
    }
    \label{fig:exp:dl3dv_comparison}
\end{figure*}

\begin{figure}[t]
    \centering
    \includegraphics[width=0.475\textwidth]{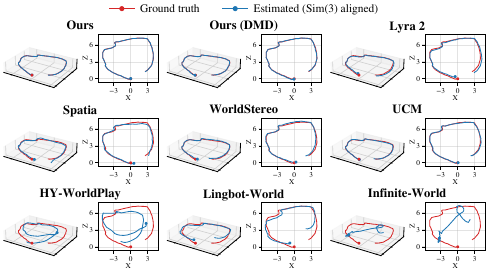}
    \caption{
    \textbf{Camera alignment.} \method achieves the best camera alignment on DL3DV-Evaluation.
    }
    \label{fig:exp:camera_align}
\end{figure}

% \vspace{-1.0 em}
\subsection{Long-Horizon Memory Mechanism}
\label{sec:method:retrieval_rerope}
As the rollout progresses, the history bank grows with the generated videos. Retaining all historical frames introduces increasing computational cost and substantial view redundancy.
We therefore organize long-term memory through coverage-based keyframe retrieval and temporal positioning, ensuring that attention accesses informative history without imposing arbitrary ordering on retrieved observations.

\vspace{-0.5 em}
\paragraph{Coverage-Based Historical Retrieval.}
We retrieve a compact history according to its geometric coverage of the upcoming target chunk.
Specifically, we project each historical frame's local geometry onto the target views, and greedily select frames that maximize newly covered regions.
Meanwhile, the initial frame and the latest frame are retained to preserve scene identity and inter-chunk continuity.
The retrieved history is thus
$
\mathcal H_t^{\mathrm{ret}}
=\mathcal H_t^{\mathrm{first}}
    \cup\mathcal H_t^{\mathrm{coverage}}
    \cup\mathcal H_t^{\mathrm{recent}}.
$

\paragraph{Order-Invariant Historical Positioning.}
Coverage key-frames are selected for geometric relevance rather than temporal proximity. Assigning their temporal RoPE positions according to concatenation order would therefore introduce artificial ordering biases into historical attention. To avoid this, we assign the same temporal index to the initial anchor and all coverage keyframes, while preserving chronological positions for recent history and target latents. With $N_C$ coverage keyframes, $N_R$ retained recent frames, and $N_T$ target latent time steps, their temporal indices are assigned as:
\begin{equation}
    \thinmuskip=2mu
    p_{\mathrm{time}}=
    [\underbrace{0,\ldots,0}_{1+N_C},
     \underbrace{1,\ldots,N_R}_{N_R},
     \underbrace{N_R{+}1,\ldots,N_R{+}N_T}_{N_T}].
    \label{eq:rope_reindexing}
\end{equation}
This makes the temporal positioning of coverage-selected memory independent of retrieval order, while preserving the temporal ordering from the recent context into the next target chunk. Together, coverage-based selection and RoPE re-indexing preserve compact, geometrically relevant history without imposing artificial temporal ordering.

\subsection{Efficient Long-Horizon Generation}
\label{sec:method:streaming_generation}

% \vspace{-2.0 em}
\paragraph{Few-step Sampling Distillation.}
To accelerate generation, we employ Distribution Matching Distillation (DMD)~\cite{yin2024onestep} to distill our attention-pruned model into a 4-step student.
We follow the standard DMD training procedure while preserving historical context conditioning, geometry-guided attention pruning, correspondence-aware attention biasing, and temporal RoPE re-indexing to maintain precise camera control and long-horizon scene consistency.

\paragraph{Streaming Memory Update.}
After each chunk, generated frames are added to history bank with camera parameters and depths estimated by a frozen 3D foundation model~\citep{depthanything3}.
This retrieve–generate–update procedure repeats across chunks, enabling long-horizon autoregressive generation.
\begin{table*}[t]
\centering
\vspace{-1.0 em}
\caption{
  \textbf{Quantitative comparison on DL3DV-Evaluation and WorldScore-Static.}
  CLIP-IQA and CLIP-Aesthetic jointly measure subjective quality; Style Cons. is reported in units of $10^{-3}$.
  Revisit metrics are reported as first / second revisit.
  \method~(DMD) denotes our four-step distilled student.
  The best and second-best results are highlighted in \textbf{bold} and \underline{underline}, respectively.
}
\label{tab:quantitative_comparison}
\begingroup
\definecolor{oursrow}{HTML}{EEF3FA}
\newcommand{\ul}[1]{\underline{#1}}
\setlength{\tabcolsep}{4.5pt}
\renewcommand{\arraystretch}{1.12}
\resizebox{\textwidth}{!}{%
\begin{tabular}{l ccc@{\hspace{14pt}}ccc@{\hspace{16pt}}ccccc@{\hspace{14pt}}c@{\hspace{14pt}}c}
\toprule
& \multicolumn{6}{c}{\textbf{DL3DV-Evaluation}}
& \multicolumn{7}{c}{\textbf{WorldScore-Static}} \\
\cmidrule(lr){2-7}\cmidrule(lr){8-14}
& \multicolumn{3}{c}{Visual Quality}
& \multicolumn{3}{c}{Camera Control}
& \multicolumn{5}{c}{Generation Quality}
& \multicolumn{2}{c}{Revisit (1st / 2nd)} \\
\cmidrule(lr){2-4}\cmidrule(lr){5-7}\cmidrule(lr){8-12}\cmidrule(lr){13-14}
\textbf{Method}
& SSIM$\uparrow$ & LPIPS$\downarrow$ & FVD$\downarrow$
& TransErr$\downarrow$ & RotErr$\downarrow$ & ATE$\downarrow$
& \makecell{Content\\Align.$\uparrow$} & \makecell{Photo.\\Cons.$\downarrow$} & \makecell{Style\\Cons.$\downarrow$}
& \makecell{CLIP\\IQA$\uparrow$} & \makecell{CLIP\\Aes.$\uparrow$}
& SSIM$\uparrow$ & LPIPS$\downarrow$ \\
\midrule
Lyra2~\cite{shen2026lyra}
& 0.3141 & 0.5280 & \textbf{704.54} & 0.0065 & \ul{0.0512} & 0.1964
& 29.23 & 0.1848 & \ul{1.58} & 0.6018 & 5.1848 & 0.4239\,/\,0.3724 & 0.3049\,/\,0.4016 \\
Spatia~\cite{zhao2026spatia}
& 0.3106 & 0.5922 & 1049.21 & 0.0395 & 0.5061 & 1.3532
& 29.14 & 0.4160 & 1.60 & 0.5723 & 5.1017 & 0.4553\,/\,0.4279 & 0.2152\,/\,0.2682 \\
HY-WorldStereo~\cite{zhang2026worldstereo}
& \ul{0.3333} & 0.6189 & 1319.87 & 0.0076 & 0.0616 & 0.3214
& 29.27 & 0.1594 & 3.12 & 0.4845 & 4.8492 & 0.4565\,/\,0.3879 & 0.4598\,/\,0.5599 \\
UCM~\cite{xu2026ucm}
& 0.3103 & 0.6406 & 1291.43 & 0.0121 & 0.1378 & 0.3571
& 29.66 & 0.1745 & 3.16 & 0.6007 & 5.2311 & 0.5423\,/\,0.5011 & 0.2095\,/\,0.2841 \\
\midrule
HY-WorldPlay~\cite{sun2025worldplay}
& 0.2002 & 0.6901 & 1649.02 & 0.0194 & 0.2770 & 1.5655
& 28.56 & 0.3421 & 10.90 & 0.5994 & 5.1965 & 0.2662\,/\,0.2014 & 0.5984\,/\,0.6915 \\
Lingbot-World~\cite{team2026advancing}
& 0.2219 & 0.7392 & 1684.33 & 0.0154 & 0.3672 & 1.2615
& 29.07 & 1.6079 & 2.62 & 0.5362 & 5.1235 & 0.2575\,/\,0.2509 & 0.7055\,/\,0.7527 \\
Infinite-World~\cite{wu2026infiniteworld}
& 0.2539 & 0.6780 & 1292.30 & 0.0345 & 0.3150 & 2.9127
& 29.66 & \textbf{0.0980} & 2.70 & 0.6328 & 5.3458 & 0.2901\,/\,0.2746 & 0.6368\,/\,0.6831 \\
\midrule
\rowcolor{oursrow}
\textbf{\method}
& \textbf{0.3352} & \textbf{0.5183} & \ul{723.62} & \textbf{0.0046} & \textbf{0.0496} & \ul{0.0903}
& \ul{30.18} & \ul{0.1080} & \textbf{1.40} & \ul{0.6790} & \ul{5.4533}
& \textbf{0.5659}\,/\,\textbf{0.5479} & \textbf{0.1800}\,/\,\textbf{0.1992} \\
\midrule
\rowcolor{oursrow}
\method~(DMD)
& 0.3236 & \ul{0.5246} & 757.58 & \ul{0.0048} & 0.0563 & \textbf{0.0760}
& \textbf{30.22} & 0.1126 & 1.61 & \textbf{0.6850} & \textbf{5.4766}
& \ul{0.5489}\,/\,\ul{0.5331} & \ul{0.2002}\,/\,\ul{0.2256} \\
\bottomrule
\end{tabular}%
}
\endgroup
\end{table*}

\section{Experiments}
\label{sec:exp}

% \vspace{-0.125 em}
\subsection{Experimental Setup}
\label{sec:exp:setup}

\paragraph{Dataset.}
We train \method on DL3DV-10K~\citep{ling2024dl3dv}, a large-scale real-world dataset with diverse camera trajectories. Each sequence is divided into 145-frame clips at a resolution of $480\times832$. 
We employ VGGT-Omega~\citep{wang2026vggtomega} to estimate camera poses and per-frame depths, and generate video captions with Qwen3-VL-8B-Instruct~\citep{bai2025qwen3}.

We construct training samples under two conditioning modes. In image-to-video (I2V) mode, we train on the first 73 frames, with the initial frame providing geometric references for target-view correspondences. In history-to-video (H2V) mode, the first 73 frames constitute the history bank, from which 4 keyframes are retrieved following the coverage-based strategy in Sec.~\ref{sec:method:retrieval_rerope} to establish token correspondence cache; the remaining 72 frames serve as generation targets.

\paragraph{Implementation Details.}
We adopt Wan2.1-I2V-14B~\citep{wan2025wan} as our pretrained backbone and implement the pruned sparse attention masks using FlexAttention~\cite{dong2024flex}.
We first perform self-attention fine-tuning with our pruning strategy for 3K iterations, followed by DMD2 distillation~\cite{yin2024improved} into a four-step student using rank-32 LoRA adapters~\cite{hu2021lora} with a CFG scale of 5 for an additional 2K iterations. Both stages are optimized using AdamW on 32 GPUs (96 GB each), with a learning rate of $1\times10^{-5}$.
\subsection{Quantitative Evaluation}
\label{sec:exp:comparison}

\paragraph{Baselines and Metrics.}
% We compare \method with recent camera-controlled video generation methods with memory mechanisms.
% Explicit-3D baselines include Lyra2~\citep{shen2026lyra}, Spatia~\citep{zhao2026spatia}, HY-WorldStereo~\citep{zhang2026worldstereo}, and UCM~\citep{xu2026ucm},
% while implicit baselines include HY-WorldPlay~\citep{sun2025worldplay}, Lingbot-World~\citep{team2026advancing}, and Infinite-World~\citep{wu2026infiniteworld}.
We compare \method against recent memory-augmented camera-controlled video generation methods, including explicit-3D approaches Lyra2~\citep{shen2026lyra}, Spatia~\citep{zhao2026spatia}, HY-WorldStereo~\citep{zhang2026worldstereo}, and UCM~\citep{xu2026ucm}, as well as implicit approaches HY-WorldPlay~\citep{sun2025worldplay}, Lingbot-World~\citep{team2026advancing}, and Infinite-World~\citep{wu2026infiniteworld}.

We first evaluate all methods on DL3DV-Evaluation~\citep{ling2024dl3dv}. Given only the same initial frame, each method autoregressively generates subsequent frames along the ground-truth camera trajectory.
% We report SSIM~\citep{wang2004image}, LPIPS~\citep{zhang2018unreasonable}, and FVD~\citep{unterthiner2018towards} to evaluate frame-level fidelity and temporal quality.
% For camera-control accuracy, we recover camera poses from the generated videos using ViPE~\citep{huang2025vipe} and align them with the ground-truth trajectories using the Umeyama transformation~\citep{umeyama1991least}. We then report TransErr, RotErr, and Absolute Trajectory Error (ATE).
We evaluate visual fidelity and temporal quality using SSIM~\citep{wang2004image}, LPIPS~\citep{zhang2018unreasonable}, and FVD~\citep{unterthiner2018towards}. To assess camera-control accuracy, we estimate camera trajectories from generated videos using ViPE~\citep{huang2025vipe}, align them with ground truth via Umeyama alignment~\citep{umeyama1991least}, and report TransErr, RotErr, and Absolute Trajectory Error (ATE).

We further evaluate long-horizon revisitation on 50 randomly sampled scenes from the WorldScore Static set~\citep{duan2025worldscore} using closed-loop camera trajectories. We report Content Alignment, Subjective Quality, Style Consistency, and Photometric Consistency, together with Revisit SSIM and Revisit LPIPS, which measure the fidelity of recovered scene content against reference observations at matched camera poses.
% which are computed between the generated revisit frame and the reference observation at the matched camera pose to assess the recovery of previously observed scene content.

\begin{figure*}[t]
    \centering
    % \vspace{-3.0 em}
    \includegraphics[width=0.975\textwidth]{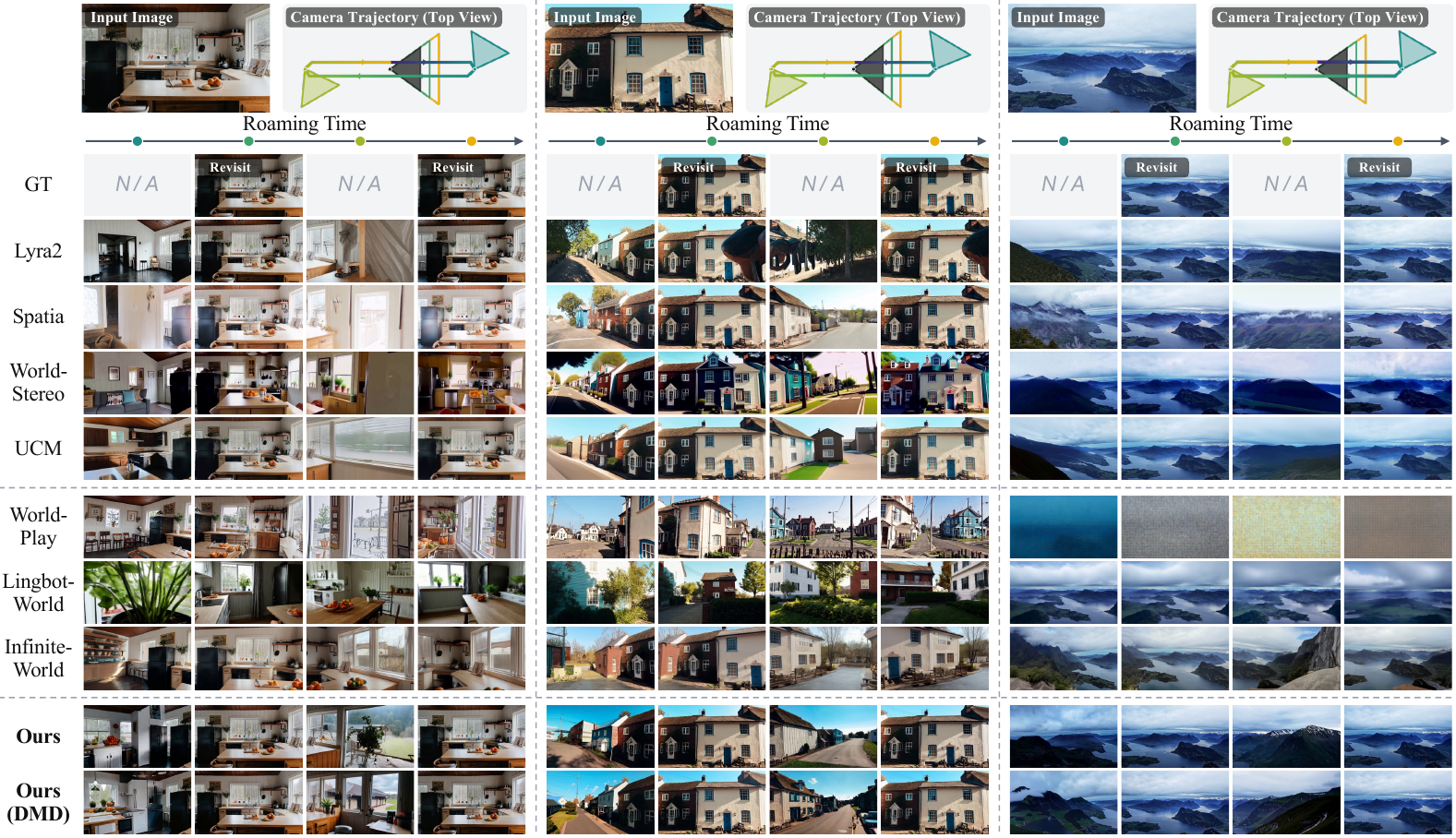}
    \caption{
    \textbf{Qualitative comparison of revisitation on WorldScore-Static.}
    Each scene follows a closed-loop camera trajectory (top view), and frames are sampled along the roaming time.
    Baselines either distort the scene or fail to recover the originally observed content upon revisiting, whereas our \method faithfully restores the scene at both revisits. \textbf{See Supp. for additional scenes and video comparisons.}
    }
    \label{fig:exp:worldscore_comparison}
\end{figure*}

\paragraph{Quantitative and Qualitative Comparison.}
As shown in Tab.~\ref{tab:quantitative_comparison}, \method performs favorably across the evaluated metrics and benchmarks.
Our method reduces ATE by 54\% compared with
the strongest baseline, demonstrating improved trajectory adherence, as further illustrated in Fig.~\ref{fig:exp:camera_align}.
% \method also achieves the best content alignment, both subjective quality metrics, and revisit performance on WorldScore, which demonstrates \method faithfully recovers previously observed content after long temporal gaps without compromising overall generation quality.
\method also achieves superior content alignment, subjective quality, and revisit performance on WorldScore, demonstrating its ability to faithfully recover observed content after long temporal gaps without compromising generation quality.

The qualitative comparisons in Figs.~\ref{fig:exp:dl3dv_comparison} and \ref{fig:exp:worldscore_comparison} further demonstrate the advantages of \method over existing methods.
% Lyra2 提供了 competitive visual fidelity, but develops camera control drift under extended camera motion.
Lyra2 delivers competitive visual fidelity but gradually drifts from the target camera path under extended camera motion.
% Spatia and WorldStereo exhibit progressive scene distortions and visual degradation, consistent with their vulnerability to accumulated reconstruction errors.
Spatia and HY-WorldStereo exhibit progressive scene distortion and visual degradation, consistent with their susceptibility to accumulated reconstruction errors in explicit 3D memory.
% UCM initially preserves coherent appearance but gradually deviates over extended rollouts.
UCM initially preserves coherent appearance but gradually deviates from the observed scene over extended rollouts.
% HY-WorldPlay and Lingbot-World struggle with both accurate camera control and revisitation, whereas Infinite-World maintains comparatively stable appearance but insufficiently follows the precise trajectory.
Among implicit methods, HY-WorldPlay and Lingbot-World struggle with both accurate camera control and revisitation, whereas Infinite-World maintains comparatively stable appearance but insufficiently follows the prescribed trajectory precisely.
% \method maintains coherent appearance and accurate camera control, 甚至相较于 baselines 对 DiT 原始架构的改动最轻微, supporting the effectiveness of 做减法反而比做加法更能体现 DiT 的潜力.
% supporting the effectiveness of separating visual memory from geometric addressing for long-horizon generation. \textbf{\textcolor{red}{Please refer to our project page} for additional scenes and more extensive video comparisons.}
In contrast, \method maintains coherent appearance and accurate camera control throughout the rollout, despite introducing the lightest architectural modification to the original DiT architecture among all compared methods.
This suggests that selectively pruning attention, rather than introducing additional memory modules, is more effective at exploiting the inherent long-horizon potential of pretrained video DiTs.
\textbf{\textcolor{red}{Please refer to supplementary material} for additional scenes and more extensive video comparisons.}
\definecolor{oursrow}{HTML}{EEF3FA}

\begin{table}[t]
\centering
\vspace{1.0 em}
\caption{
  \textbf{Ablation on attention pruning.}
  Geometric correspondence attention, local window attention, and correspondence bias provide complementary improvements in reconstruction fidelity, while shared historical RoPE further enhances generation quality.
}
\label{tab:ablation_attention}
\begingroup
\newcommand{\ul}[1]{\underline{#1}}
\newcommand{\hd}[1]{{\scriptsize #1}}
\footnotesize
\setlength{\tabcolsep}{1.4pt}
\renewcommand{\arraystretch}{1.08}
\begin{tabular*}{\columnwidth}{@{\extracolsep{\fill}}l cccc cccc@{}}
  \toprule
  \textbf{Variant}
  & \hd{\makecell{Hist.\\Corr.}} & \hd{\makecell{Local\\Win.}} & \hd{\makecell{Corr.\\Bias}} & \hd{\makecell{Shared\\RoPE}}
  & \hd{PSNR$\uparrow$} & \hd{SSIM$\uparrow$} & \hd{LPIPS$\downarrow$} & \hd{ATE$\downarrow$} \\
  \midrule
  \makecell[l]{Dense Attn.\\\quad + Cam.\ Emb.}
  & \xmark & \xmark & \xmark & \cmark & 12.06 & 0.287 & 0.519 & 0.279 \\
  \makecell[l]{Dense Attn.\\\quad + Corr.\ Bias.}
  & \xmark & \xmark & \cmark & \cmark & 16.01 & 0.414 & 0.274 & 0.039 \\
  \midrule
  Hist.\ Corr.\ Only
  & \cmark & \xmark & \xmark & \cmark & 14.53 & 0.339 & 0.472 & 0.360 \\
  + Local Window
  & \cmark & \cmark & \xmark & \cmark & 15.36 & 0.407 & 0.295 & 0.041 \\
  \rowcolor{oursrow}
  + Corr.\ Bias (\textbf{Ours})
  & \cmark & \cmark & \cmark & \cmark & \textbf{16.80} & \textbf{0.473} & \textbf{0.232} & \textbf{0.018} \\
  \midrule
  Ours w/ Seq.\ RoPE
  & \cmark & \cmark & \cmark & \xmark & \ul{16.38} & \ul{0.454} & \ul{0.244} & \ul{0.022} \\
  \bottomrule
\end{tabular*}
\endgroup
\end{table}

\begin{figure}[t]
    \centering
    \vspace{-0.5 em}
    \includegraphics[width=\columnwidth]{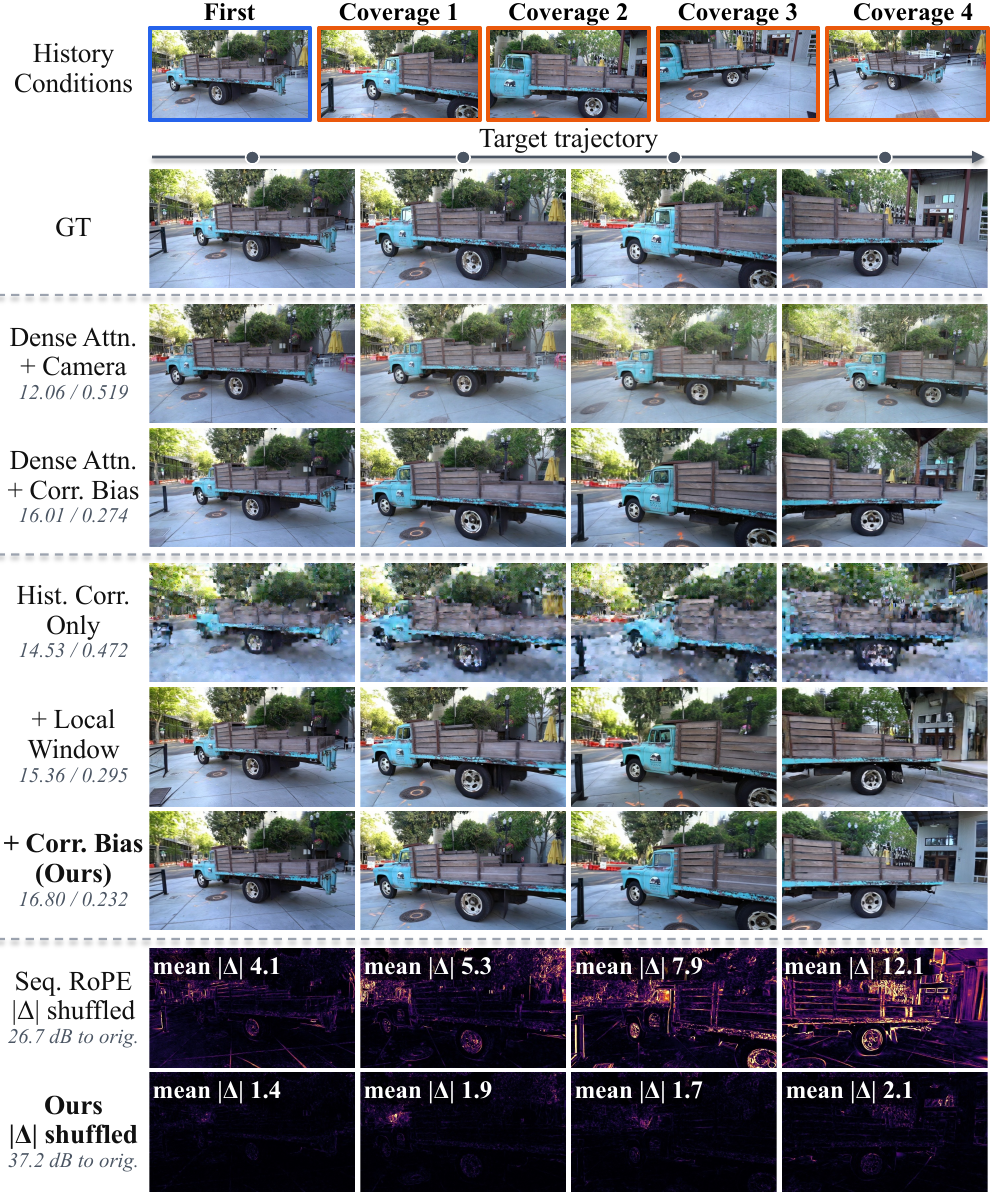}
    \vspace{-1.0 em}
    \caption{
    \textbf{Ablation on attention pruning and historical RoPE indexing.} Progressively incorporating geometric correspondence, local window attention, and correspondence bias improves camera adherence and reconstruction fidelity. 
    % Shared historical RoPE further reduces appearance variations under shuffled history order, as illustrated by the pixel-wise difference maps.
    The bottom rows visualize the pixel-wise difference $|\Delta|$ between outputs generated using the original and shuffled history orders:
    sequential history RoPE yields discrepancies that grow along the trajectory, whereas our shared indexing remains invariant.
    }
    \label{fig:exp:ablation_attention}
    \vspace{-0.25 em}
\end{figure}

\subsection{Ablation Study}
\label{sec:exp:ablation}

To investigate the contributions of individual design choices to generation quality and computational efficiency, we conduct controlled novel-view generation experiments on Tanks and Temples Dataset~\cite{knapitsch2017tanks}. 
We sample camera-calibrated multi-view images as historical observations and generate videos along held-out camera trajectories with ground-truth frames available for evaluation. All variants are trained with identical data and optimization settings following Sec.~\ref{sec:exp:setup}.
% For fair comparisons, we train all variants with identical training data, learning rates, and training iterations.

% figure* floats to the top of the next page, \ie, the Conclusion page

\begin{figure}[t]
    \centering
    \includegraphics[width=\columnwidth]{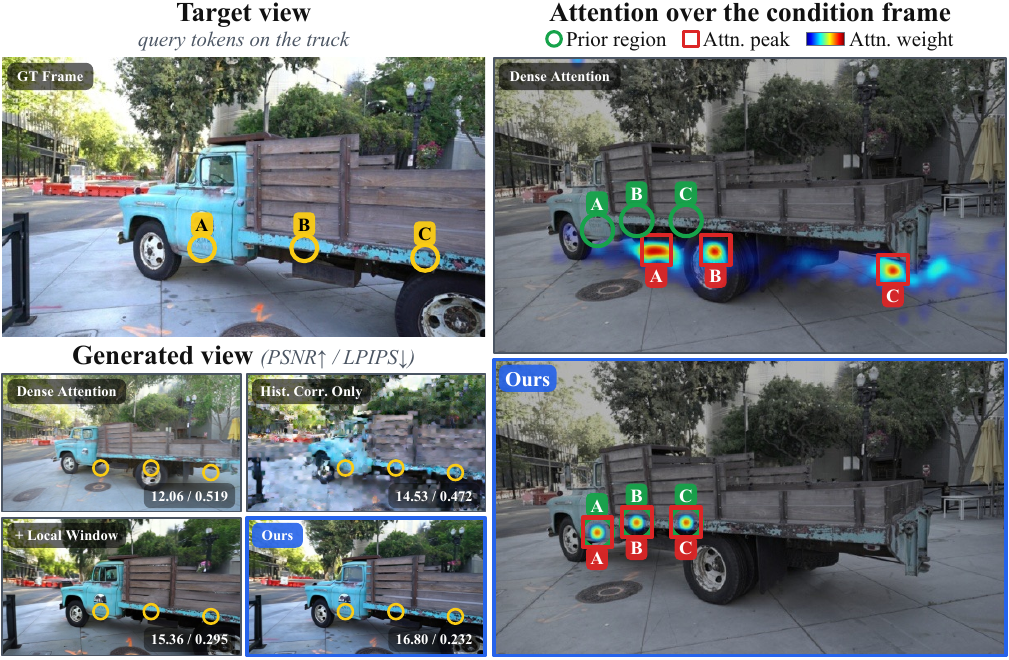}
    \caption{
    \textbf{Attention weight visualization.} Given target queries, dense attention with the camera adapter exhibits attention peaks that deviate from geometrically corresponding historical regions.
    }
    \label{fig:exp:attention_weight_vis}
\end{figure}

\paragraph{Attention Pruning Components.}
We first establish a baseline that retains the pretrained DiT's dense self-attention and incorporates camera conditioning through the camera adapter following~\cite{he2024cameractrl}. Despite producing relative viewpoint changes, it exhibits poor camera adherence and progressive visual degradation, yielding the lowest fidelity in Tab.~\ref{tab:ablation_attention} and Fig.~\ref{fig:exp:ablation_attention}. 
% We further visualize its historical attention distributions in Fig.~\ref{fig:exp:attention_weight_vis}. Dense attention exhibits pronounced spatial misalignment, with attention peaks deviating from geometric correspondences and only 0.4--1.3\% of historical attention mass assigned to matched regions for the illustrated queries.
As illustrated in Fig.~\ref{fig:exp:attention_weight_vis}, its historical attention peaks frequently deviate from geometrically corresponding regions, with only 0.4–-1.3\% of the historical attention mass assigned to matched regions for the illustrated queries. 
Introducing correspondence bias into dense attention substantially improves camera adherence and reconstruction fidelity (Fig.~\ref{fig:exp:ablation_attention}), revealing that accurate camera control can emerge from a sparse subset of spatially aligned attention interactions. However, its inferior performance to our pruned counterpart suggests that prioritizing relevant interactions alone cannot fully eliminate interference from unrelated tokens.

We then progressively examine our pruning strategy. 
Restricting observed target queries only to geometrically corresponding historical tokens ($\mathcal{N}_{\mathrm{hist}}$) enables camera-guided generation and memory retrieval directly through attention pruning, but produces mosaic-like artifacts due to insufficient local context.
% However, attending to only a handful of matched tokens produces noticeable mosaic-like artifacts and texture discontinuities due to insufficient local context. 
Introducing local window attention ($\mathcal{N}_{\mathrm{local}}$) restores spatiotemporal coherence and substantially improves perceptual quality, while correspondence bias further strengthens geometrically matched historical evidence against the substantially larger local target context.
These findings demonstrate that geometry-guided pruning does more than reduce computation: it restructures information flow to suppress irrelevant interactions and strengthen spatially aligned historical evidence, providing a unified mechanism for camera control and persistent scene memory.

\paragraph{Order-Invariant Historical Positioning.}
Since coverage-based retrieval selects keyframes by geometric relevance, their concatenation order carries no meaningful temporal information in long-horizon rollouts, and the generation should therefore be invariant to it.
To verify this, we run inference under four permutations of the retrieved history, and evaluate 
our shared historical RoPE against conventional sequential RoPE.
Permuting the retrieved historical frames introduces substantial appearance variations under sequential RoPE, whereas our shared indexing produces nearly identical outputs, as illustrated in Fig.~\ref{fig:exp:ablation_attention} and Tab.~\ref{tab:ablation_rope}. 
% This confirms the importance of eliminating artificial temporal ordering biases among geometrically retrieved historical observations during long-horizon rollout.
This highlights the importance of eliminating spurious positional biases toward temporally adjacent historical frames to enable permutation-invariant memory retrieval.

%%%%%%%%%%%%%% Efficiency Evaluation %%%%%%%%%%%%%%
\paragraph{Computational Analysis.}
We profile the computational cost of each design in Tab.~\ref{tab:ablation_efficiency}.
Each variant generates a 72-frame chunk at $480\times832$ conditioned on 9 historical latents, \ie, 42,120 tokens in total, of which 99.0\% of target tokens are observed.
All measurements are conducted on a single 96 GB GPU.
Restricting observed queries to historical correspondences skips 98.1\% of query--key pairs, reducing attention FLOPs by $53\times$ and per-step latency by $2.70\times$.
Adding local window attention with $w=2$ enlarges this scope but still skips 86.7\% of query--key pairs.
Distilling the pruned model into a 4-step student removes classifier-free guidance and reduces function evaluations from 100 to 4, 
achieving a $53.6\times$ speedup in per-chunk DiT inference.
% shortening DiT time per chunk with $53.6\times$.

\definecolor{oursrow}{HTML}{EEF3FA}

\begin{table}[t]
\centering
\vspace{0.875 em}
\caption{
  \textbf{Robustness to historical frame ordering.}
  Compared with sequential RoPE, our shared historical RoPE reduces output variations under shuffled history order. Results are reported as mean ± standard deviation.
}
\label{tab:ablation_rope}
\begingroup
\newcommand{\std}[1]{\,{\tiny$\pm$#1}}
\footnotesize
\setlength{\tabcolsep}{2.8pt}
\renewcommand{\arraystretch}{1.08}
\begin{tabular}{@{}l ccc cc@{}}
  \toprule
  & \multicolumn{3}{c}{vs.\ GT (mean\,$\pm$\,std)} & \multicolumn{2}{c}{vs.\ Original Order} \\
  \cmidrule(lr){2-4}\cmidrule(l){5-6}
  \textbf{History RoPE}
  & PSNR$\uparrow$ & SSIM$\uparrow$ & LPIPS$\downarrow$ & PSNR$\uparrow$ & LPIPS$\downarrow$ \\
  \midrule
  Sequential
  & 16.38\std{0.02} & 0.455\std{0.001} & 0.246\std{0.002} & 26.67 & 0.051 \\
  \rowcolor{oursrow}
  Shared (\textbf{Ours})
  & \textbf{16.79}\std{0.00} & \textbf{0.473}\std{0.000} & \textbf{0.232}\std{0.000} & \textbf{37.21} & \textbf{0.007} \\
  \bottomrule
\end{tabular}
\endgroup
\end{table}

% \subsection{Training-Free Dynamic Scene Generation}
% \label{sec:exp:dynamic_generation}
\paragraph{Training-Free Dynamic Scene Generation.}

\begin{figure}[t]
    \centering
    \includegraphics[width=\columnwidth]{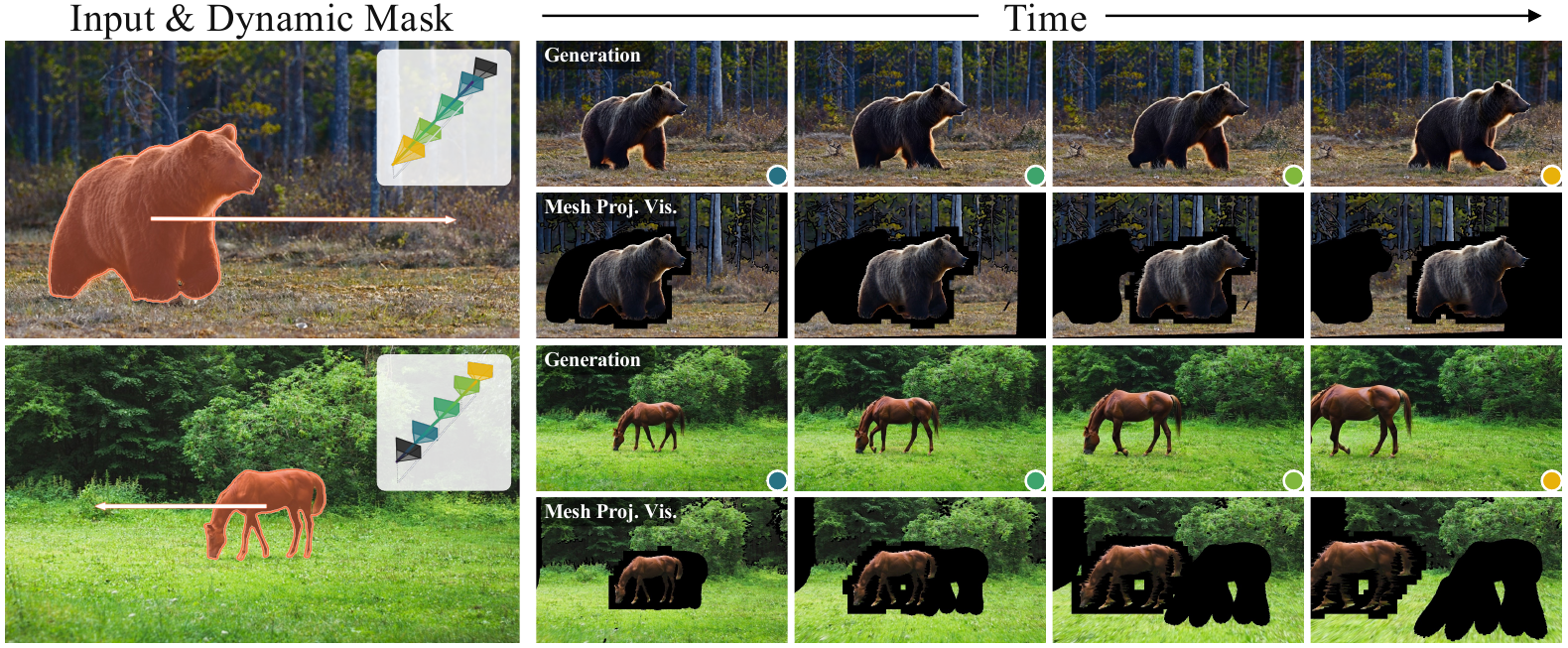}
    \caption{
    \textbf{Training-free dynamic scene generation.}
    We segment the foreground object and shift its correspondence locations along a prescribed motion trajectory, allowing \method to synthesize natural object motion consistent with the specified trajectory.
    }
    \label{fig:exp:dynamic_generation}
\end{figure}

Although trained exclusively on static scenes, \method naturally extends to dynamic scene generation without additional training. Since geometric correspondences govern the scope of historical attention, manipulating these correspondences enables controllable object motion while preserving background consistency. Specifically, as shown in Fig.~\ref{fig:exp:dynamic_generation}, we segment the foreground object and shift its correspondence locations along a prescribed motion trajectory, allowing \method to synthesize natural object motion consistent with the specified trajectory. These results highlight the potential of correspondence manipulation for joint camera and object motion control in a training-free manner.

\begin{figure*}[t]
    \centering
    % \vspace{-1.0 em}
    \includegraphics[width=\textwidth]{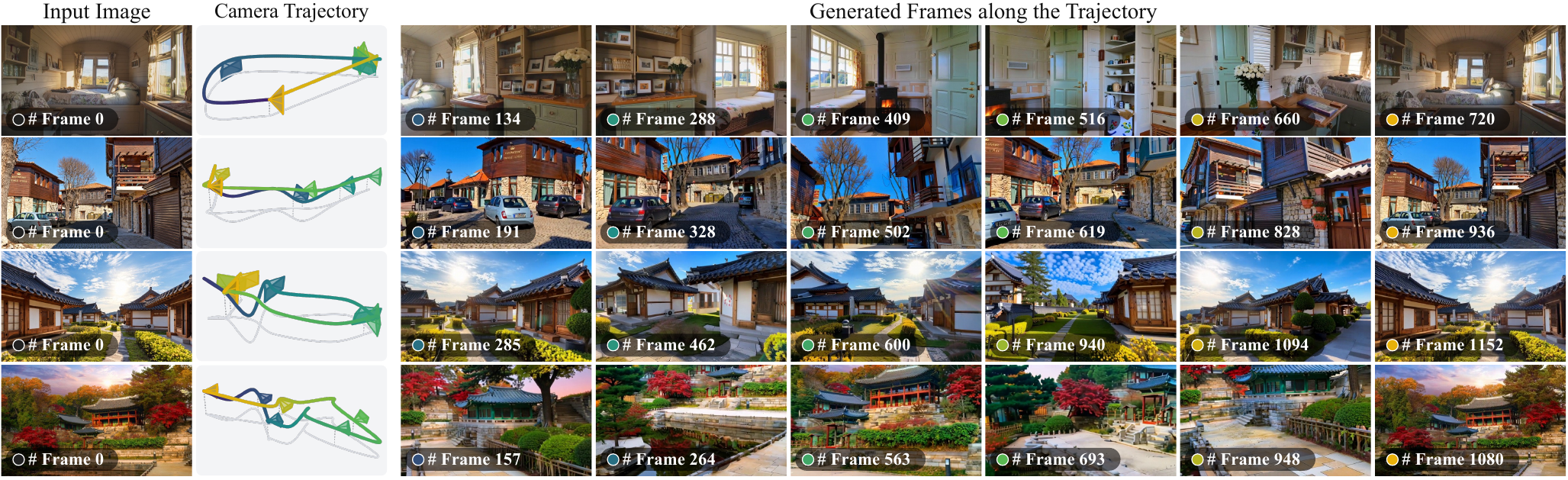}
    \caption{
    \textbf{Additional long-horizon generation results.}
    Without any degradation-aware augmentation, \method generates high-fidelity, minute-long videos with precise camera adherence and persistent scene consistency.
    % without 任何防画面退化的augmentation，SPARC generates minute-long visual fidelity videos while faithfully following the camera motion and maintaining long-horizon scene appearance consistency.
    }
    \label{fig:more_results}
\end{figure*}

\begin{table}[t]
\centering
\vspace{1.25 em}
\caption{
  \textbf{Computational efficiency.}
  By combining attention pruning with 4-step distillation, \method achieves a \textbf{50$\times$ inference speedup} over the original dense backbone.
}
\label{tab:ablation_efficiency}
\begingroup
\definecolor{oursrow}{HTML}{EEF3FA}
\newcommand{\ul}[1]{\underline{#1}}
\newcommand{\pt}[1]{\,{\tiny(#1)}}
\footnotesize
\setlength{\tabcolsep}{1.5pt}
\renewcommand{\arraystretch}{1.08}
\begin{tabular*}{\columnwidth}{@{\extracolsep{\fill}}l cc cc cc@{}}
\toprule
& & & Token & TFLOPs & Per Chunk & \\
\textbf{Variant}
& Steps & CFG & Sparsity & /step & Time (s) & Speedup \\
\midrule
Dense (Wan2.1)
& 50 & \cmark & \phantom{0}0.0\% & 4980\pt{2907} & 2436.1 & 1.0$\times$ \\
Hist.\ Corr.\ Only
& 50 & \cmark & 98.1\% & \ul{2129}\pt{55} & \ul{903.6} & \ul{2.7$\times$} \\
\quad + Local Win.
& 50 & \cmark & 86.7\% & 2461\pt{387} & 1090.4 & 2.2$\times$ \\
\rowcolor{oursrow}
\quad + DMD (\textbf{Ours})
& \phantom{0}4 & \xmark & 86.7\% & \textbf{1241}\pt{194} & \textbf{45.5} & \textbf{53.6$\times$} \\
\bottomrule
\end{tabular*}
\endgroup
\end{table}

% \vspace{-0.5 em}
\section{Conclusion}
\label{sec:conclusion}
% \vspace{-0.25 em}

We present \method, a geometry-guided attention pruning framework that unifies camera control and persistent scene memory by restructuring the native self-attention of pretrained video diffusion transformers. Our key insight is that attention connectivity itself can serve as geometric conditioning, and that selectively removing geometrically irrelevant interactions improves spatial control, revisit consistency, and computational efficiency. Without additional learnable control or memory modules, \method enables efficient minute-long camera-controlled video generation. Our findings highlights attention connectivity as a promising interface for unlocking the inherent spatial reasoning and memory capabilities of pretrained video diffusion models.

\paragraph{Limitations.} \method relies on external geometry estimation to establish spatial correspondences, introducing additional computational overhead. 
A promising direction is to jointly learn geometry reasoning and attention connectivity within the generative model, enabling a more unified and efficient framework.

{
    \small
    % \nocite{*} % TODO: remove once the paper has real \cite commands; otherwise every main.bib entry is listed
    \bibliographystyle{ieeenat_fullname}
    \bibliography{main}

@article{wan2025wan,
  title   = {{Wan}: Open and advanced large-scale video generative models},
  author  = {{Team Wan} and Wang, Ang and Ai, Baole and Wen, Bin and Mao, Chaojie
             and Xie, Chen-Wei and Chen, Di and Yu, Feiwu and Zhao, Haiming
             and Yang, Jianxiao and Zeng, Jianyuan and others},
  journal = {arXiv preprint arXiv:2503.20314},
  year    = {2025},
  doi     = {10.48550/arXiv.2503.20314},
  url     = {https://arxiv.org/abs/2503.20314}
}

@article{seedance2026seedance2,
  title   = {{Seedance 2.0}: Advancing video generation for world complexity},
  author  = {{Team Seedance}},
  journal = {arXiv preprint arXiv:2604.14148},
  year    = {2026},
  doi     = {10.48550/arXiv.2604.14148},
  url     = {https://arxiv.org/abs/2604.14148}
}

@misc{minimax2026h3,
  title        = {MiniMax H3: An Open Model Breaking the Boundaries Between Tasks and Modalities},
  author       = {{MiniMax Research}},
  year         = {2026},
  month        = {July},
  howpublished = {\url{https://www.minimax.io/blog/minimax-h3}},
  note         = {Official Technical Blog}
}

@misc{klingai2026kling3,
  title        = {{Kling VIDEO 3.0} model user guide},
  author       = {{Kling AI}},
  year         = {2026},
  month        = feb,
  howpublished = {\url{https://app.klingai.com/global/quickstart/klingai-video-3-model-user-guide}},
  note         = {Official model documentation}
}

@misc{googledeepmind2025veo3,
  title        = {{Veo 3} model card},
  author       = {{Google DeepMind}},
  year         = {2025},
  month        = may,
  howpublished = {\url{https://storage.googleapis.com/deepmind-media/Model-Cards/Veo-3-Model-Card.pdf}},
  note         = {Published May 23, 2025; updated January 13, 2026}
}

@article{bai2025qwen3,
  title={{Qwen3-VL} technical report},
  author={Bai, Shuai and Cai, Yuxuan and Chen, Ruizhe and Chen, Keqin and Chen, Xionghui and Cheng, Zesen and Deng, Lianghao and Ding, Wei and Gao, Chang and Ge, Chunjiang and others},
  journal={arXiv preprint arXiv:2511.21631},
  year={2025}
}

@inproceedings{he2024cameractrl,
  title={{CameraCtrl}: Enabling Camera Control for Text-to-Video Generation},
  author={He, Hao and Xu, Yinghao and Guo, Yuwei and Wetzstein, Gordon and Dai, Bo and Li, Hongsheng and Yang, Ceyuan},
  booktitle={International Conference on Learning Representations},
  year={2025}
}

@inproceedings{bai2025recammaster,
  title={{ReCamMaster}: Camera-controlled generative rendering from a single video},
  author={Bai, Jianhong and Xia, Menghan and Fu, Xiao and Wang, Xintao and Mu, Lianrui and Cao, Jinwen and Liu, Zuozhu and Hu, Haoji and Bai, Xiang and Wan, Pengfei and Zhang, Di},
  booktitle={Proceedings of the IEEE/CVF International Conference on Computer Vision},
  pages={14834--14844},
  year={2025}
}

@misc{li2025hunyuangamecrafthighdynamicinteractivegame,
  title={{Hunyuan-GameCraft}: High-dynamic interactive game video generation with hybrid history condition},
  author={Li, Jiaqi and Tang, Junshu and Xu, Zhiyong and Wu, Longhuang and Zhou, Yuan and Shao, Shuai and Yu, Tianbao and Cao, Zhiguo and Lu, Qinglin},
  year={2025},
  eprint={2506.17201},
  archivePrefix={arXiv},
  primaryClass={cs.CV},
  url={https://arxiv.org/abs/2506.17201}
}

@article{team2026advancing,
  title={Advancing open-source world models},
  author={{Robbyant Team} and Gao, Zelin and Wang, Qiuyu and Zeng, Yanhong and Zhu, Jiapeng and Cheng, Ka Leong and Li, Yixuan and Wang, Hanlin and Xu, Yinghao and Ma, Shuailei and others},
  journal={arXiv preprint arXiv:2601.20540},
  year={2026}
}

@misc{alibaba2026happyoyster,
  title        = {{Happy Oyster}: An open-ended world model for real-time world creation and interaction},
  author       = {{Alibaba Token Hub}},
  year         = {2026},
  month        = apr,
  howpublished = {\url{https://www.happyoyster.com/}},
  note         = {Accessed: 2026-08-28}
}

@misc{alibaba2026happyhorse,
  title        = {{Happy Horse}: Text-to-Video Generation},
  author       = {{Alibaba Token Hub}},
  year         = {2026},
  month        = apr,
  howpublished = {\url{https://www.happyhorse.com/}},
  note         = {Accessed: 2026-10-28}
}

@misc{parkerholder2025genie3,
  title        = {{Genie 3}: A new frontier for world models},
  author       = {Parker-Holder, Jack and Fruchter, Shlomi},
  year         = {2025},
  month        = aug,
  howpublished = {\url{https://deepmind.google/blog/genie-3-a-new-frontier-for-world-models/}},
  note         = {Google DeepMind; accessed 2026-08-28}
}

@article{chen2026hy,
  title={Hy-Nomas: Towards Free-form Interactive World Generation},
  author={Chen, Qiaoling and Cheng, Jiaxiang and Deng, Xueqing and Hang, Tiankai and Hou, Ji and Li, Jiaqi and Liu, Yifan and Lu, Tao and Tang, Junshu and Tang, Yu and others},
  journal={arXiv preprint arXiv:2610.36729},
  year={2026}
}

@article{lingbot-world-v2,
      title={Infinite Worlds with Versatile Interactions}, 
      author={Zelin Gao and Qiuyu Wang and Jiapeng Zhu and Jingye Chen and Zichen Liu and Qingyan Bai and Jiahao Wang and Yufeng Yuan and Hanlin Wang and Yichong Lu and Ka Leong Cheng and Haojie Zhang and Jian Gao and Tianrui Feng and Yuzheng Liu and Yao Yao and Yinghao Xu and Xing Zhu and Yujun Shen and Hao Ouyang},
      journal={arXiv preprint arXiv:2607.07534},
      year={2026}
}

@misc{zhang2026echowmopenenterableomnimodal,
  title={EchoWM: Open and Enterable Omnimodal World Models},
  author={Songchun Zhang and Yaowei Li and Junhao Zhuang and Weiyang Jin and Haoyu Wang and Xin Lu and Yilang Sun and Shiyi Zhang and Haoran Li and Xiaoxiao Ma and Yuming Li and Yijun Liu and Yaofeng Su and Yanwen Ma and Haoyu Wu and Zihan Su and Yue Ma and Lvmin Zhang and Haoyang Huang and Zeyue Xue and Anyi Rao and Nan Duan},
  year={2026},
  eprint={2608.23189},
  archivePrefix={arXiv},
  primaryClass={cs.CV},
  url={https://arxiv.org/abs/2608.23189},
}

@inproceedings{mao2025yume15textcontrolledinteractiveworld,
  title     = {{Yume1.5}: A Text-Controlled Interactive World Generation Model},
  author    = {Mao, Xiaofeng and Li, Zhen and Li, Chuanhao and Xu, Xiaojie and Ying, Kaining and Zhang, Kaipeng},
  booktitle = {Proceedings of the IEEE/CVF Conference on Computer Vision and Pattern Recognition},
  year      = {2026}
}

@inproceedings{xiao2026worldmem,
  title={{WorldMem}: Long-term consistent world simulation with memory},
  author={Xiao, Zeqi and Lan, Yushi and Zhou, Yifan and Ouyang, Wenqi and Yang, Shuai and Zeng, Yanhong and Pan, Xingang},
  booktitle={Advances in Neural Information Processing Systems},
  pages={49632--49652},
  year={2025}
}

@inproceedings{li2025vmem,
  title={{VMem}: Consistent interactive video scene generation with surfel-indexed view memory},
  author={Li, Runjia and Torr, Philip and Vedaldi, Andrea and Jakab, Tomas},
  booktitle={Proceedings of the IEEE/CVF International Conference on Computer Vision},
  pages={25690--25699},
  year={2025}
}

@inproceedings{yu2025cam,
  title={{Context as Memory}: Scene-consistent interactive long video generation with memory retrieval},
  author={Yu, Jiwen and Bai, Jianhong and Qin, Yiran and Liu, Quande and Wang, Xintao and Wan, Pengfei and Zhang, Di and Liu, Xihui},
  booktitle={SIGGRAPH Asia 2025 Conference Papers},
  publisher={ACM},
  year={2025},
  doi={10.1145/3757377.3763833}
}

@inproceedings{sun2025worldplay,
  title     = {{WorldPlay}: Towards Long-Term Geometric Consistency for Real-Time Interactive World Modeling},
  author    = {Sun, Wenqiang and Zhang, Haiyu and Wang, Haoyuan and Wu, Junta and Wang, Zehan and Wang, Zhenwei and Wang, Yunhong and Zhang, Jun and Wang, Tengfei and Guo, Chunchao},
  booktitle = {International Conference on Machine Learning},
  year      = {2026}
}

@article{hong2025relic,
  title   = {{RELIC}: Interactive video world model with long-horizon memory},
  author  = {Hong, Yicong and Mei, Yiqun and Ge, Chongjian and Xu, Yiran and Zhou, Yang and Bi, Sai and Hold-Geoffroy, Yannick and Roberts, Mike and Fisher, Matthew and Shechtman, Eli and Sunkavalli, Kalyan and Liu, Feng and Li, Zhengqi and Tan, Hao},
  journal = {arXiv preprint arXiv:2512.04040},
  year    = {2025}
}

@inproceedings{wu2026infiniteworld,
  title={{Infinite-World}: Scaling interactive world models to 1000-frame horizons via pose-free hierarchical memory},
  author={Wu, Ruiqi and He, Xuanhua and Cheng, Meng and Yang, Tianyu and Zhang, Yong and Kang, Zhuoliang and Cai, Xunliang and Wei, Xiaoming and Guo, Chunle and Li, Chongyi and Cheng, Ming-Ming},
  booktitle={International Conference on Machine Learning},
  year={2026}
}

@inproceedings{zhang2025framepack,
  title={Frame context packing and drift prevention in next-frame-prediction video diffusion models},
  author={Zhang, Lvmin and Cai, Shengqu and Li, Muyang and Wetzstein, Gordon and Agrawala, Maneesh},
  booktitle={Advances in Neural Information Processing Systems},
  year={2025}
}

@inproceedings{ren2025gen3c,
  title={{GEN3C}: {3D}-informed world-consistent video generation with precise camera control},
  author={Ren, Xuanchi and Shen, Tianchang and Huang, Jiahui and Ling, Huan and
      Lu, Yifan and Nimier-David, Merlin and M{\"u}ller, Thomas and Keller, Alexander and
      Fidler, Sanja and Gao, Jun},
  booktitle={Proceedings of the IEEE/CVF Conference on Computer Vision and Pattern Recognition},
  year={2025}
}

@inproceedings{wu2025video,
  title={Video world models with long-term spatial memory},
  author={Wu, Tong and Yang, Shuai and Po, Ryan and Xu, Yinghao and Liu, Ziwei and Lin, Dahua and Wetzstein, Gordon},
  booktitle={Advances in Neural Information Processing Systems},
  year={2025}
}

@inproceedings{zhao2026spatia,
  title={{Spatia}: Video generation with updatable spatial memory},
  author={Zhao, Jinjing and Wei, Fangyun and Liu, Zhening and Zhang, Hongyang and Xu, Chang and Lu, Yan},
  booktitle={Proceedings of the IEEE/CVF Conference on Computer Vision and Pattern Recognition},
  pages={4245--4257},
  year={2026}
}

@inproceedings{zhang2026worldstereo,
  title={{WorldStereo}: Bridging camera-guided video generation and scene reconstruction via {3D} geometric memories},
  author={Zhang, Yisu and Cao, Chenjie and Wang, Tengfei and Zuo, Xuhui and Wu, Junta and Zhu, Jianke and Guo, Chunchao},
  booktitle={Proceedings of the IEEE/CVF Conference on Computer Vision and Pattern Recognition},
  pages={40327--40339},
  year={2026}
}

@inproceedings{lee20253dscenepromptingsceneconsistent,
  title     = {{3D} Scene Prompting for Scene-Consistent Camera-Controllable Video Generation},
  author    = {Lee, JoungBin and Jung, Jaewoo and Han, Jisang and Narihira, Takuya and Fukuda, Kazumi and Seo, Junyoung and Hong, Sunghwan and Mitsufuji, Yuki and Kim, Seungryong},
  booktitle = {International Conference on Learning Representations},
  year      = {2026}
}

@inproceedings{wang2026anchorweave,
  title={{AnchorWeave}: World-consistent video generation with retrieved local spatial memories},
  author={Wang, Zun and Lin, Han and Yoon, Jaehong and Cho, Jaemin and Zhang, Yue and Bansal, Mohit},
  booktitle={European Conference on Computer Vision},
  organization={Springer},
  year={2026}
}

@inproceedings{chen2026anyrecon,
  title     = {{AnyRecon}: Arbitrary-View {3D} Reconstruction with Video Diffusion Model},
  author    = {Chen, Yutian and Guo, Shi and Jin, Renbiao and Yang, Tianshuo and Cai, Xin and Luo, Yawen and Yang, Mingxin and Yu, Mulin and Xu, Linning and Xue, Tianfan},
  booktitle = {ACM SIGGRAPH Asia 2026 Conference Papers},
  publisher = {ACM},
  year      = {2026}
}

@article{shen2026lyra,
  title={{Lyra 2.0}: Explorable generative {3D} worlds},
  author={Shen, Tianchang and Bahmani, Sherwin and He, Kai and Srinivasan, Sangeetha Grama and Cao, Tianshi and Ren, Jiawei and Li, Ruilong and Wang, Zian and Sharp, Nicholas and Gojcic, Zan and others},
  journal={arXiv preprint arXiv:2604.13036},
  year={2026}
}

@inproceedings{xu2026ucm,
  title={{UCM}: Unified modeling of camera control and memory with time-aware positional encoding warping for world models},
  author={Xu, Tianxing and Wang, Zixuan and Wang, Guangyuan and Hu, Li and Zhang, Zhongyi and Zhang, Peng and Zhang, Bang and Zhang, Songhai},
  booktitle={ACM SIGGRAPH 2026 Conference Papers},
  publisher={ACM},
  year={2026},
  doi={10.1145/3799902.3811088}
}

@article{liu2024omnidirector,
  title={OmniDirector: General Multi-Shot Camera Cloning without Cross-Paired Data},
  author={Liu, Jiwen and Li, Shujuan and Fang, Zhixue and Li, Xiaohan and Zhou, Yan and Meng, Zijie and Zhang, Zhimin and Luo, Yawen and Zhang, Guoxin and Wan, Pengfei},
  journal={arXiv preprint arXiv:2606.13432},
  year={2026}
}

@inproceedings{yin2024onestep,
      title={One-step Diffusion with Distribution Matching Distillation},
      author={Yin, Tianwei and Gharbi, Micha{\"e}l and Zhang, Richard and Shechtman, Eli and Durand, Fr{\'e}do and Freeman, William T and Park, Taesung},
      booktitle={CVPR},
      year={2024}
    }

@article{xu2026wonder,
  title={Wonder: Video World Model Done Better},
  author={Xu, Jiacong and Jiang, Hanwen and Shu, Zhixin and Sunkavalli, Kalyan and Patel, Vishal M. and Mei, Yiqun},
  journal={arXiv preprint arXiv:2607.26037},
  year={2026},
  url={https://arxiv.org/abs/2607.26037}
}

@misc{hu2025ex4dextremeviewpoint4d,
      title={EX-4D: EXtreme Viewpoint 4D Video Synthesis via Depth Watertight Mesh}, 
      author={Hu, Tao and Peng, Haoyang and Liu, Xiao and Ma, Yuewen},
      year={2025},
      eprint={2506.05554},
      archivePrefix={arXiv},
      primaryClass={cs.CV},
      url={https://arxiv.org/abs/2506.05554}
}

@inproceedings{hu2021lora,
  title={{LoRA}: Low-rank adaptation of large language models},
  author={Hu, Edward J and Shen, Yelong and Wallis, Phillip and Allen-Zhu, Zeyuan and Li, Yuanzhi and Wang, Shean and Wang, Lu and Chen, Weizhu},
  booktitle={International Conference on Learning Representations},
  year={2022}
}

@inproceedings{lipman2022flow,
  title={Flow matching for generative modeling},
  author={Lipman, Yaron and Chen, Ricky TQ and Ben-Hamu, Heli and Nickel, Maximilian and Le, Matt},
  booktitle={International Conference on Learning Representations},
  year={2023}
}

@inproceedings{depthanything3,
  title={{Depth Anything 3}: Recovering the visual space from any views},
  author={Lin, Haotong and Chen, Sili and Liew, Jun Hao and Chen, Donny Y. and Li, Zhenyu and Shi, Guang and Feng, Jiashi and Kang, Bingyi},
  booktitle={International Conference on Learning Representations},
  year={2026}
}

@article{huang2025vipe,
  title={{ViPE}: Video pose engine for {3D} geometric perception},
  author={Huang, Jiahui and Zhou, Qunjie and Rabeti, Hesam and Korovko, Aleksandr and Ling, Huan and Ren, Xuanchi and Shen, Tianchang and Gao, Jun and Slepichev, Dmitry and Lin, Chen-Hsuan and others},
  journal={arXiv preprint arXiv:2508.10934},
  year={2025}
}

@inproceedings{ling2024dl3dv,
  title={{DL3DV-10K}: A large-scale scene dataset for deep learning-based {3D} vision},
  author={Ling, Lu and Sheng, Yichen and Tu, Zhi and Zhao, Wentian and Xin, Cheng and Wan, Kun and Yu, Lantao and Guo, Qianyu and Yu, Zixun and Lu, Yawen and others},
  booktitle={Proceedings of the IEEE/CVF Conference on Computer Vision and Pattern Recognition},
  pages={22160--22169},
  year={2024}
}

@inproceedings{duan2025worldscore,
  title={{WorldScore}: A unified evaluation benchmark for world generation},
  author={Duan, Haoyi and Yu, Hong-Xing and Chen, Sirui and Fei-Fei, Li and Wu, Jiajun},
  booktitle={Proceedings of the IEEE/CVF International Conference on Computer Vision},
  pages={27713--27724},
  year={2025}
}

@article{wang2004image,
  title={Image quality assessment: from error visibility to structural similarity},
  author={Wang, Zhou and Bovik, Alan C and Sheikh, Hamid R and Simoncelli, Eero P},
  journal={IEEE Transactions on Image Processing},
  volume={13},
  number={4},
  pages={600--612},
  year={2004},
  publisher={IEEE}
}

@inproceedings{zhang2018unreasonable,
  title={The unreasonable effectiveness of deep features as a perceptual metric},
  author={Zhang, Richard and Isola, Phillip and Efros, Alexei A and Shechtman, Eli and Wang, Oliver},
  booktitle={Proceedings of the IEEE/CVF Conference on Computer Vision and Pattern Recognition},
  pages={586--595},
  year={2018}
}

@article{unterthiner2018towards,
  title={Towards accurate generative models of video: A new metric \& challenges},
  author={Unterthiner, Thomas and Van Steenkiste, Sjoerd and Kurach, Karol and Marinier, Raphael and Michalski, Marcin and Gelly, Sylvain},
  journal={arXiv preprint arXiv:1812.01717},
  year={2018}
}

@article{umeyama1991least,
  title={Least-squares estimation of transformation parameters between two point patterns},
  author={Umeyama, Shinji},
  journal={IEEE Transactions on Pattern Analysis and Machine Intelligence},
  volume={13},
  number={4},
  pages={376--380},
  year={1991},
  publisher={IEEE}
}

@article{bai2025positional,
  title={Positional Encoding Field},
  author={Bai, Yunpeng and Li, Haoxiang and Huang, Qixing},
  journal={arXiv preprint arXiv:2510.20385},
  year={2025}
}

@article{knapitsch2017tanks,
  title={{Tanks and Temples}: Benchmarking Large-Scale Scene Reconstruction},
  author={Knapitsch, Arno and Park, Jaesik and Zhou, Qian-Yi and Koltun, Vladlen},
  journal={ACM Transactions on Graphics},
  volume={36},
  number={4},
  pages={78:1--78:13},
  year={2017}
}

@article{li2025cameras,
    title={Cameras as Relative Positional Encoding},
    author={Li, Ruilong and Yi, Brent and Liu, Junchen and Gao, Hang and Ma, Yi and Kanazawa, Angjoo},
    journal={Advances in Neural Information Processing Systems},
    year={2025}
}

@article{yin2026scope,
  title={SCoPE: Sightline-Coordinate Positional Encoding for Video Diffusion Transformers},
  author={Yin, Minghao and Lu, Jiahao and Hu, Wenbo and Zhao, Wang and Shan, Ying and Han, Kai},
  year={2026}
}

@article{yu2024viewcrafter,
  title={ViewCrafter: Taming Video Diffusion Models for High-fidelity Novel View Synthesis},
  author={Yu, Wangbo and Xing, Jinbo and Yuan, Li and Hu, Wenbo and Li, Xiaoyu and Huang, Zhipeng and Gao, Xiangjun and Wong, Tien-Tsin and Shan, Ying and Tian, Yonghong},
  journal={arXiv preprint arXiv:2409.02048},
  year={2024}
}

@inproceedings{zhang2026unified,
  title={Unified camera positional encoding for controlled video generation},
  author={Zhang, Cheng and Li, Boying and Wei, Meng and Cao, Yan-Pei and Gambardella, Camilo and Phung, Dinh and Cai, Jianfei},
  booktitle={Proceedings of the IEEE/CVF Conference on Computer Vision and Pattern Recognition},
  pages={38027--38037},
  year={2026}
}

@article{zhou2025stable,
    title={Stable Virtual Camera: Generative View Synthesis with Diffusion Models},
    author={Jensen (Jinghao) Zhou and Hang Gao and Vikram Voleti and Aaryaman Vasishta and Chun-Han Yao and Mark Boss and
    Philip Torr and Christian Rupprecht and Varun Jampani
    },
    journal={arXiv preprint},
    year={2025}
}

@inproceedings{wang2026vggtomega,
  title     = {{VGGT-$\Omega$}},
  author    = {Jianyuan Wang and Minghao Chen and Shangzhan Zhang and Nikita Karaev and Johannes Sch{\"o}nberger and Patrick Labatut and Piotr Bojanowski and David Novotny and Andrea Vedaldi and Christian Rupprecht},
  booktitle = {Proceedings of the IEEE/CVF Conference on Computer Vision and Pattern Recognition (CVPR)},
  year      = {2026}
}

@article{dong2024flex,
  title={Flex attention: A programming model for generating optimized attention kernels},
  author={Dong, Juechu and Feng, Boyuan and Guessous, Driss and Liang, Yanbo and He, Horace},
  journal={arXiv preprint arXiv:2412.05496},
  volume={2},
  number={3},
  pages={4},
  year={2024}
}

@inproceedings{yin2024improved,
    title={Improved Distribution Matching Distillation for Fast Image Synthesis},
    author={Yin, Tianwei and Gharbi, Micha{\"e}l and Park, Taesung and Zhang, Richard and Shechtman, Eli and Durand, Fredo and Freeman, William T},
    booktitle={NeurIPS},
    year={2024}
}

@article{yang2026geometry,
    title={Geometry as Address: Routing Attention to Visual Memory for Long-Horizon Camera-Controlled Video Generation},
    author={Yang, Zesong and Chen, Weikai and Cui, Liyuan and Jiang, Lutao and Zhang, Runze and Yin, Yingda and Huang, Xiaoyang and Yan, Kai and Luo, Keyang and Zheng, Wangguandong and Wang, Xin and Bao, Hujun and Cui, Zhaopeng},
    journal={arXiv preprint arXiv:2609.34722},
    year={2026}
}
}

% The supplementary material is compiled separately from supp.tex.

\end{document}